\documentclass[10pt,twocolumn,letterpaper]{article}

\usepackage{titling}
\usepackage{iccv}
\usepackage{times}
\usepackage{epsfig}
\usepackage{graphicx}
\usepackage{amsmath}
\usepackage{amssymb}
\usepackage{booktabs}
\usepackage{multirow}
\usepackage{wrapfig}
\usepackage[table]{xcolor}
\usepackage{color}
\usepackage{xspace}
\usepackage{hhline}
\usepackage{cite}
\usepackage{microtype}
\usepackage{pifont}
\usepackage{caption}
\usepackage{algorithm} 
\usepackage{algpseudocode} 
\usepackage{makecell} 

\usepackage[pagebackref=true,breaklinks=true,letterpaper=true,colorlinks,bookmarks=false]{hyperref}

\usepackage[capitalize]{cleveref}
\crefname{section}{Sec.}{Secs.}
\Crefname{section}{Section}{Sections}
\Crefname{table}{Table}{Tables}
\crefname{table}{Tab.}{Tabs.}

\iccvfinalcopy 

\def\iccvPaperID{0000} 

\definecolor{darkblue}{rgb}{0, 0.2, 0.6}
\definecolor{gray}{rgb}{0.2, 0.2, 0.2}
\definecolor{lightgray}{rgb}{0.45, 0.45, 0.45}
\definecolor{deepdarkblue}{rgb}{0, 0.05, 0.3}
\definecolor{orange}{rgb}{1.0, 0.5, 0.0}
\definecolor{darkorange}{rgb}{0.7, 0.3, 0.0}
\definecolor{red}{rgb}{1.0, 0.0, 0.0}
\definecolor{pastelred}{rgb}{0.9, 0.2, 0.45}
\definecolor{purple}{rgb}{0.6, 0.0, 0.6}
\definecolor{dogwoodrose}{rgb}{0.84, 0.09, 0.41}
\definecolor{mint}{rgb}{0.01176, 0.5490, 0.5490}
\definecolor{blue}{rgb}{0, 0, 1.0}
\definecolor{azure(colorwheel)}{rgb}{0.0, 0.5, 1.0}
\definecolor{nicegreen}{rgb}{0.0, 0.7, 0.1}
\definecolor{CuGray}{gray}{0.9}
\definecolor{amethyst}{rgb}{0.6, 0.4, 0.8}
\definecolor{black}{rgb}{0.0, 0.0, 0.0}
\definecolor{steelblue}{rgb}{0.27, 0.51, 0.7}
\definecolor{brightcerulean}{rgb}{0.11, 0.67, 0.84}
\definecolor{brown}{rgb}{0.4, 0.3, 0.3}

\newcommand{\redcolornumber}[1]{{\color{red}{#1}}}

\newcommand{\firststyle}[1]{{\color{pastelred}{#1}}}

\newcommand{\classname}[1]{{\color{gray}{#1}}}
\newcommand{\commentcolor}[1]{{\color{lightgray}{#1}}}

\makeatletter
\DeclareRobustCommand\onedot{\futurelet\@let@token\@onedot}
\def\@onedot{\ifx\@let@token.\else.\null\fi\xspace}
\def\eg{\emph{e.g}\onedot} 
\def\ie{\emph{i.e}\onedot}

\newcommand*\bigcdot{\mathpalette\bigcdot@{.5}}
\newcommand*\bigcdot@[2]{\mathbin{\vcenter{\hbox{\scalebox{#2}{$\m@th#1\bullet$}}}}}
\makeatother

\begin{document}

\title{PromptStyler: Prompt-driven Style Generation \\
for Source-free Domain Generalization}
\author{Junhyeong Cho$^1$ \quad Gilhyun Nam$^1$ \quad Sungyeon Kim$^2$ \quad Hunmin Yang$^{1,3}$ \quad Suha Kwak$^2$\\
{\small $^1$ADD \qquad $^2$POSTECH \qquad $^3$KAIST}\\
{\small \url{https://PromptStyler.github.io}}}
\date{}

\maketitle 
\begin{abstract}
\vspace{-3mm}
In a joint vision-language space, a text feature (\eg, from ``a photo of a dog") could effectively represent its relevant image features (\eg, from dog photos). 
Also, a recent study has demonstrated the cross-modal transferability phenomenon of this joint space.
From these observations,
we propose \textbf{PromptStyler} which simulates various distribution shifts in the joint space by synthesizing diverse styles via prompts without using any images to deal with source-free domain generalization.
The proposed method learns to generate a variety of style features (from ``a $\boldsymbol{S_{*}}$ style of a") via learnable style word vectors for pseudo-words $\boldsymbol{S_{*}}$. 
To ensure that learned styles do not distort content information, 
we force \mbox{style-content} features (from ``a $\boldsymbol{S_{*}}$ style of a [class]") to be located nearby their corresponding content features (from ``[class]") in the joint vision-language space. 
After learning style word vectors, we train a linear classifier using synthesized \mbox{style-content} features.
PromptStyler achieves the state of the art on PACS, VLCS, OfficeHome and DomainNet, 
even though it does not require any images for training.
\end{abstract}
\vspace{-2mm}
\section{Introduction}
\vspace{-0.5mm}

\begin{figure}[t!]
    \centering
    \includegraphics[width=\columnwidth]{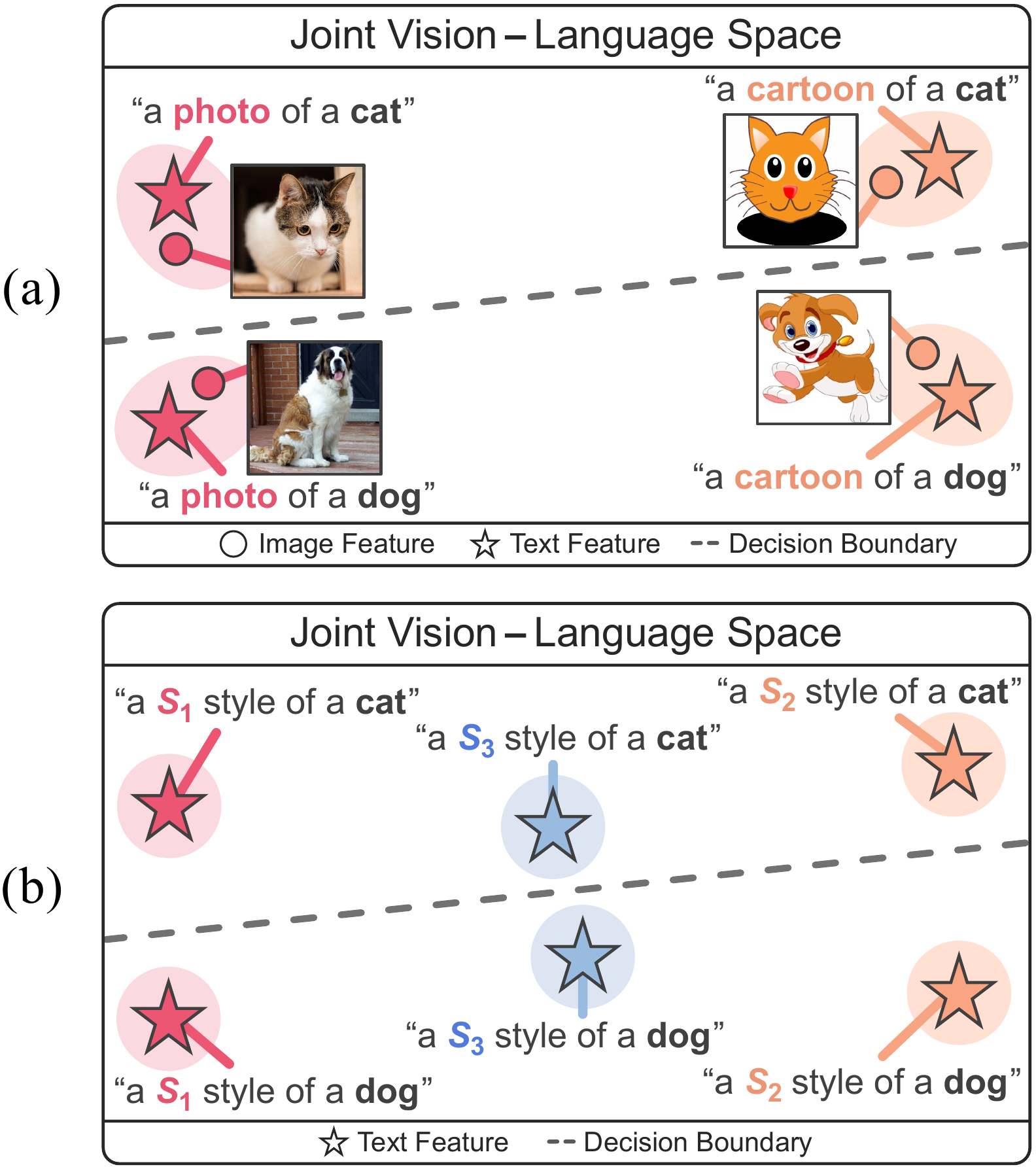}
    \vspace{-5.7mm}
    \caption{Motivation of our method.
    (a) Text features could effectively represent various image styles in a joint vision-language space.
    (b) PromptStyler synthesizes diverse styles in a joint vision-language space via learnable style word vectors for pseudo-words $\boldsymbol{S_{*}}$ without using any images.}
    \label{fig:fig1}
    \vspace{-1.2mm}
\end{figure}

Deep neural networks are usually trained with the assumption that training and test data are independent and identically distributed,
which makes them vulnerable to substantial distribution shifts between training and test data~\cite{recht2019ood,hendrycks2019ood}.
This susceptibility is considered as one of the major obstacles to their deployment in real-world applications.
To enhance their robustness to such distribution shifts,
Domain Adaptation (DA)~\cite{ben2006da,sun2016coral,hoffman2018cycada,tzeng2017adversarial,lee2022fifo,zhao2019da,Saito2019semi,lee2023surgical} has been studied; 
it aims at adapting neural networks to a target domain using target domain data available in training.
However, such a target domain is often latent in common training scenarios, which considerably limits the application of DA.
Recently, a body of research has addressed this limitation by Domain Generalization (DG)~\cite{Li2018Domain,zhou2021mixstyle,gulrajani2021domainbed,Li2018Learning,Carlucci2019Domain,cha2022miro,kim2022broad} that aims to improve model's generalization capability to any unseen domains.
It has been a common practice in DG to utilize multiple source domains for learning domain-invariant features~\cite{zhou2022dgsurvey,wang2021dgsurvey}, 
but it is unclear which source domains are ideal for DG, 
since arbitrary unseen domains should be addressed.
Furthermore, it is costly and sometimes even infeasible to collect and annotate large-scale multi-source domain data for training.

\begin{figure*}[t!]
    \centering
    \includegraphics[width=\textwidth]{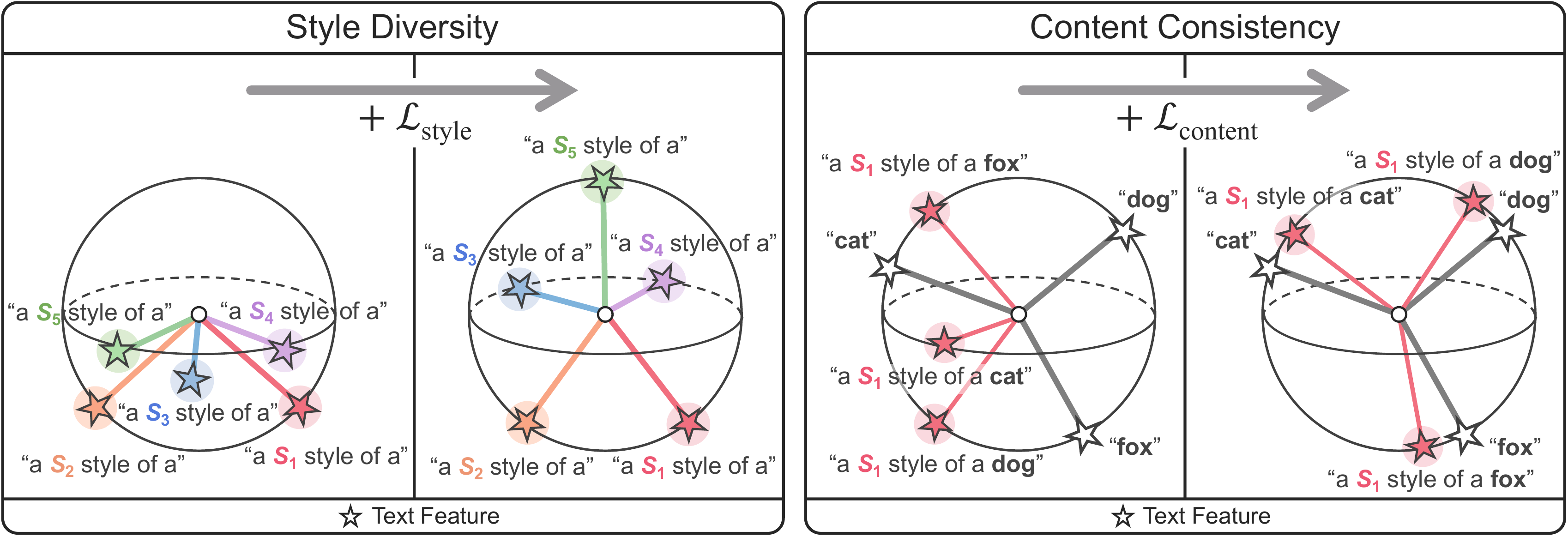}
    \vspace{-6.4mm}
    \caption{Important factors in the proposed method.
    PromptStyler learns style word vectors for 
    pseudo-words $\boldsymbol{S_{*}}$ which lead to diverse style features (from ``a $\boldsymbol{S_{*}}$ style of a") while preserving content information encoded in \mbox{style-content} features (from ``a $\boldsymbol{S_{*}}$ style of a [class]").
    $\mathcal{L}_{\mathrm{style}}$ and $\mathcal{L}_{\mathrm{content}}$ are the loss functions used for maximizing \textit{style diversity} and \textit{content consistency} in a hyperspherical joint vision-language space (\eg, CLIP~\cite{radford2021clip} latent space).}
    \vspace{-2.3mm}
    \label{fig:fig2}
\end{figure*}

We notice that a large-scale pre-trained model might have already observed a great variety of domains and thus can be used as an efficient proxy of actual multiple source domains.
From this perspective, we raised a question
\textit{``Could we further improve model's generalization capability by simulating various distribution shifts in the latent space of such a large-scale model without using any source domain data?"}
If this is possible, DG will become immensely practical by effectively and efficiently exploiting such a large-scale model. 
However, this approach is much more challenging since any actual data of source and target domains are not accessible but only the target task definition (\eg, class names) is given.

In this paper, we argue that large-scale vision-language models~\cite{ALIGN,radford2021clip,TCL} could shed light on this challenging \textit{source-free domain generalization}.
As conceptually illustrated in Figure~\ref{fig:fig1}(a),
text features could effectively represent their relevant image features in a joint vision-language space.
Despite the modality gap between two modalities in the joint space~\cite{liang2022mind},
a recent study has demonstrated the cross-modal transferability phenomenon~\cite{zhang2023diagnosing};
we could train a classifier using text features while running an inference with the classifier using image features.
This training procedure meets the necessary condition for the source-free domain generalization, 
\ie, source domain images are not required.
Using such a joint vision-language space, we could simulate various distribution shifts via prompts without any images.

We propose a prompt-driven style generation method, dubbed \textbf{PromptStyler}, which synthesizes diverse styles via learnable word vectors to simulate distribution shifts in a hyperspherical joint vision-language space.
PromptStyler is motivated by the observation that a shared style of images could characterize a domain~\cite{zhou2021mixstyle,kang2022styleneophile} and such a shared style could be captured by a learnable word vector for a pseudo-word $\boldsymbol{S_{*}}$ using CLIP~\cite{radford2021clip} with a prompt (``a painting in the style of $\boldsymbol{S_{*}}$")~\cite{gal2023textualinversion}.
As shown in Figure~\ref{fig:fig1}(b), our method learns a style word vector for $\boldsymbol{S_{*}}$ to represent each style.

To effectively simulate various distribution shifts, we try to maximize \textit{style diversity} as illustrated in Figure~\ref{fig:fig2}.
Specifically, our method encourages learnable style word vectors to result in orthogonal style features in the hyperspherical space,
where each style feature is obtained from a \textbf{style prompt} (``a $\boldsymbol{S_{*}}$ style of a") via a pre-trained text encoder.
To prevent learned styles from distorting content information, we also consider \textit{content consistency} as illustrated in Figure~\ref{fig:fig2}.
Each style-content feature obtained from a \textbf{\mbox{style-content} prompt} (``a $\boldsymbol{S_{*}}$ style of a [class]") is forced to be located closer to its corresponding content feature obtained from a \textbf{content prompt} (``[class]") than the other content features.

Learned style word vectors are used to synthesize style-content features for training a classifier; 
these synthesized features could simulate images of known contents with diverse unknown styles in the joint space.
These style-content features are fed as input to a linear classifier which is trained by a classification loss using contents (``[class]") as their class labels. 
At inference time, an image encoder extracts image features from input images, which are fed as input to the trained classifier.
Note that the text and image encoders are derived from the same pre-trained vision-language model (\eg, CLIP~\cite{radford2021clip}); the text encoder is only involved in training and the image encoder is only involved at inference time.

The proposed method achieves state-of-the-art results on PACS~\cite{PACSdataset}, VLCS~\cite{VLCSdataset}, OfficeHome~\cite{OfficeHomedataset} and DomainNet~\cite{DomainNetdataset} without using any actual data of source and target domains.
It takes just $\sim$30 minutes for the entire training using a single RTX 3090 GPU, and our model is $\sim$2.6$\times$ smaller and $\sim$243$\times$ faster at inference compared with CLIP~\cite{radford2021clip}.

\begin{table}[!t]
    \centering
    \resizebox{\columnwidth}{!}{
        \begin{tabular}{cccc}
        \Xhline{2\arrayrulewidth}
        Setup & Source
        & Target & Task Definition
        \\
        \hline
            DA & \ding{51}
            & \ding{51} & \ding{51} 
        \\
            DG & \ding{51}
            & \textbf{--}  & \ding{51} 
        \\
            Source-free DA & \textbf{--} 
            & \ding{51} & \ding{51} 
        \\
            \cellcolor{gray!9.0}\textbf{Source-free DG} & \cellcolor{gray!9.0}\textbf{--} 
            & \cellcolor{gray!9.0}\textbf{--} & \cellcolor{gray!9.0}\ding{51} 
        \\
        \Xhline{2\arrayrulewidth}
    \end{tabular}}
    \vspace{-2mm}
    \caption{
        Different requirements in each setup.
        Source-free DG only assumes the task definition (\ie, what should be predicted) without requiring source and target domain data.
    }
    \vspace{-2.5mm}
    \label{table:task_definition}
\end{table}

Our contributions are summarized as follows:
\begin{itemize}
    \vspace{-2.5mm}
    \item This work is the first attempt to synthesize a variety of styles in a joint vision-language space via prompts to effectively tackle source-free domain generalization.
    \vspace{-2.5mm}
    \item This paper proposes a novel method that effectively simulates images of known contents with diverse unknown styles in a joint vision-language space. 
    \vspace{-2.5mm}
    \item PromptStyler achieves the state of the art on domain generalization benchmarks without using any images.
\end{itemize}
\section{Related Work}

\noindent \textbf{Domain Generalization.}
Model's generalization capability to arbitrary unseen domains is the key factor to successful deployment of neural networks in real-world applications,
since substantial distribution shifts between source and target domains could significantly degrade their performance ~\cite{recht2019ood,hendrycks2019ood}.
To this end, Domain Generalization (DG)~\cite{Muandet2013DG,Li2018Learning,Li2018Domain,gulrajani2021domainbed,wang2021dgsurvey,zhou2022dgsurvey,Min2022Grounding,SWAD,kim2022broad,LRDG2022,cha2022miro,frikha2022data} has been studied.
It assumes target domain data are not accessible while using data from source domains.
Generally speaking, existing DG methods could be divided into two categories: multi-source DG~\cite{Zhou2020Learning,Li2019Episodic,Carlucci2019Domain,Dou2019Domain,Matsuura2020Domain,Seo2020Learning,Mahajan2021Domain,zhou2021mixstyle,Xu2021FACT,Rame2022Fishr} and single-source DG~\cite{Wang2021Learning,Li2021Progressive,Qiao2020Learning,Fan2021Adversarially}.
Mostly, multi-source DG methods aim to learn domain-invariant features by exploiting available multiple source domains, and single-source DG methods also aim to learn such features by generating diverse domains based on a single domain and then exploiting the synthesized domains.

\noindent \textbf{Source-free Domain Generalization.}
In this setup, we are not able to access any source and target domains as summarized in Table~\ref{table:task_definition}.
Thus, source-free DG is much more challenging than multi-source and single-source DG. 
From the observation that 
synthesizing new domains from the given source domain could effectively improve model's generalization capability~\cite{Zhou2020Learning,zhou2020ddaig,Wang2021Learning,Li2021Progressive,kang2022styleneophile},
we also try to generate diverse domains but without using any source domains to deal with source-free DG.
By leveraging a large-scale pre-trained model which has already seen a great variety of domains,
our method could simulate various distribution shifts in the latent space of the large-scale model.
This approach has several advantages compared with existing DG methods;
source domain images are not required and there is no concern for catastrophic forgetting which might impede model's generalization capability.
Also, it would be immensely practical to exploit such a large-scale model for downstream visual recognition tasks, since we only need the task definition.

\noindent \textbf{Large-scale model in Domain Generalization.}
Recently, several DG methods~\cite{cha2022miro,CAD} exploit a large-scale pre-trained model (\eg, CLIP~\cite{radford2021clip}) to leverage its great generalization capability.
While training neural networks on available data, CAD~\cite{CAD} and MIRO~\cite{cha2022miro}
try to learn robust features using such a large-scale model.
Compared with them, the proposed method could learn domain-invariant features using a large-scale pre-trained model without requiring any actual data.

\noindent \textbf{Joint vision-language space.}
Large-scale vision-language models~\cite{ALIGN,radford2021clip,TCL} are trained with a great amount of image-text pairs, and achieve state-of-the-art results on downstream visual recognition tasks~\cite{CoOp,CoCoOp,ProDA,CLIPAdapter,TipAdapter}.
By leveraging their joint vision-language spaces, we could also effectively manipulate visual features via prompts~\cite{StyleGAN_NADA,StlyeCLIP,CLIPstyler,LADS}. 
Interestingly, Textual Inversion~\cite{gal2023textualinversion} shows that a learnable style word vector for a pseudo-word $\boldsymbol{S_{*}}$ could capture a shared style of images using CLIP~\cite{radford2021clip} with a prompt (``a painting in the style of $\boldsymbol{S_{*}}$").
From this observation, we argue that learnable style word vectors would be able to seek a variety of styles for simulating various distribution shifts in a joint vision-language space without using any images.
\section{Method}

\begin{figure*}[t!]
    \centering
    \includegraphics[width=\textwidth]{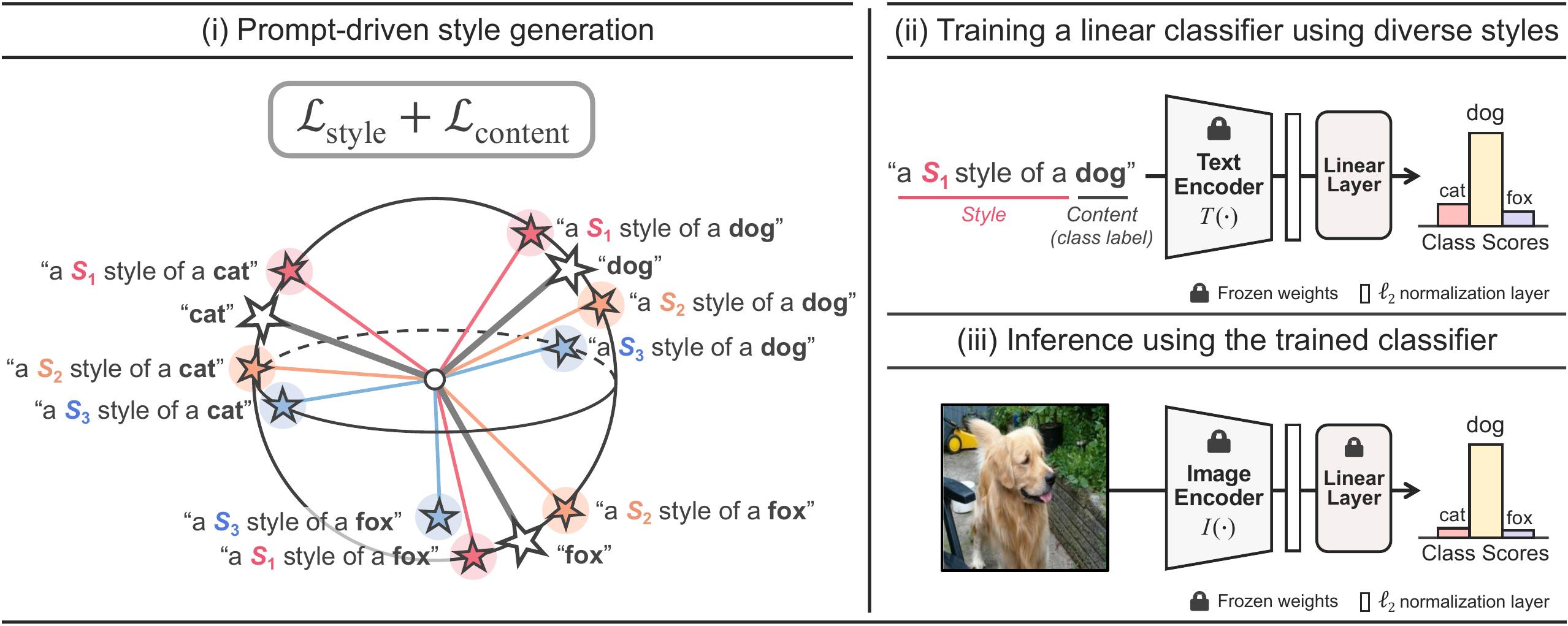}
    \vspace{-6mm}
    \caption{\mbox{PromptStyler} learns diverse style word vectors which do not distort content information of style-content prompts.
    After learning style word vectors, we synthesize \mbox{style-content} features (\eg, from ``a $\boldsymbol{\firststyle{S_{1}}}$ style of a \classname{\textbf{dog}}") via a pre-trained text encoder for training a linear classifier.
    The classifier is trained by a classification loss using those synthesized features and their corresponding class labels (\eg, ``\classname{\textbf{dog}}").
    At inference time, a pre-trained image encoder extracts image features, which are fed as input to the trained classifier. 
    Note that the encoders are derived from the same vision-language model (\eg, CLIP~\cite{radford2021clip}).}
    \vspace{-1.8mm}
    \label{fig:fig3}
\end{figure*}

The overall framework of the proposed method is shown in Figure~\ref{fig:fig3}, and pseudo-code of PromptStyler is described in Algorithm~\ref{alg:pseudocode_prompt}.
Our method learns style word vectors to represent a variety of styles in a hyperspherical joint vision-language space (\eg, CLIP~\cite{radford2021clip} latent space).
After learning those style word vectors, we train a linear classifier using synthesized style-content features produced by a pre-trained text encoder $T(\cdot)$.
At inference time, a pre-trained image encoder $I(\cdot)$ extracts image features from input images, which are fed as input to the trained linear classifier.
Thanks to the cross-modal transferability phenomenon of the joint vision-language space~\cite{zhang2023diagnosing},
this classifier could produce class scores using the image features.
Note that we exploit CLIP as our large-scale vision-language model;
its image encoder and text encoder are frozen in our entire framework.

\subsection{Prompt-driven style generation}
\label{main:main_3_1}
An input text prompt is converted to several tokens via a tokenization process, and then such tokens are replaced by their corresponding word vectors via a word lookup process.
In \mbox{PromptStyler}, a pseudo-word $\boldsymbol{S}_{i}$ in a prompt is a placeholder which is replaced by a style word vector $\mathbf{s}_{i} \in \mathbb R^{D}$ during the word lookup process.
Note that three kinds of prompts are used in the proposed method:
a style prompt $\mathcal{P}_{i}^{\,\mathrm{style}}$ (``a $\boldsymbol{S}_{i}$ style of a"), a content prompt $\mathcal{P}_{m}^{\,\mathrm{content}}$ (``[class]$_{m}$"), and a style-content prompt $\mathcal{P}_{i}^{\,\mathrm{style}} \circ \mathcal{P}_{m}^{\,\mathrm{content}}$ (``a $\boldsymbol{S}_{i}$ style of a [class]$_{m}"$).
$\boldsymbol{S}_{i}$ indicates the placeholder for $i$-th style word vector and [class]$_{m}$ denotes $m$-th class name.

Suppose we want to generate $K$ different styles in a joint vision-language space.
In this case, the proposed method needs to learn $K$ style word vectors $\{\mathbf{s}_{i}\}^{K}_{i=1}$, where each $\mathbf{s}_{i}$ is randomly initialized at the beginning.
To effectively simulate various distribution shifts in the joint vision-language space,
those style word vectors need to be diverse while not distorting content information when they are exploited in style-content prompts.
There are two possible design choices for learning such word vectors: 
(1) learning each style word vector $\mathbf{s}_{i}$ in a sequential manner, or (2) learning all style word vectors $\{\mathbf{s}_{i}\}^{K}_{i=1}$ in a parallel manner.
We choose the former, since it takes much less memory during training.
Please refer to the supplementary material (Section \redcolornumber{A.2}) for the empirical justification of our design choice.

\noindent \textbf{Style diversity loss.}
To maximize the diversity of $K$ styles in a hyperspherical joint vision-language space,
we sequentially learn style word vectors $\{\mathbf{s}_{i}\}^{K}_{i=1}$ in such a way that $i$-th style feature $T(\mathcal{P}_{i}^{\,\mathrm{style}}) \in \mathbb R^{C}$ produced by $i$-th style word vector  $\mathbf{s}_{i}$ is orthogonal to $\{T(\mathcal{P}_{j}^{\,\mathrm{style}})\}^{i-1}_{j=1}$ produced by previously learned style word vectors $\{\mathbf{s}_{j}\}^{i-1}_{j=1}$.
Regarding this, the style diversity loss $\mathcal{L}_{\mathrm{style}}$ for learning $i$-th style word vector $\mathbf{s}_{i}$ is computed by
\begin{align}
    \label{eq:loss_style}
    \mathcal{L}_{\mathrm{style}} &= \frac{1}{i-1} \sum^{i-1}_{j=1} \left| \frac{T(\mathcal{P}_{i}^{\,\mathrm{style}})}{\Vert T(\mathcal{P}_{i}^{\,\mathrm{style}}) \Vert_{\scriptscriptstyle{2}}} \bigcdot \frac{T(\mathcal{P}_{j}^{\,\mathrm{style}})}{\Vert T(\mathcal{P}_{j}^{\,\mathrm{style}}) \Vert_{\scriptscriptstyle{2}}} \right| \;.
\end{align}
This style loss $\mathcal{L}_{\mathrm{style}}$ aims to minimize the absolute value of the cosine similarity between $i$-th style feature and each of the existing style features.
When the value of this loss becomes zero, it satisfies the orthogonality between $i$-th style feature and all the existing style features.

\noindent \textbf{Content consistency loss.}
Learning the style word vectors $\{\mathbf{s}_{i}\}^{K}_{i=1}$ only using the style diversity loss sometimes leads to undesirable outcome, 
since a learned style $\mathbf{s}_{i}$ could substantially distort content information when used to generate a style-content feature $T(\mathcal{P}_{i}^{\,\mathrm{style}} \circ \mathcal{P}_{m}^{\,\mathrm{content}}) \in \mathbb R^{C}$.
To alleviate this problem, we encourage the content information in the style-content feature to be consistent with its corresponding content feature $T(\mathcal{P}_{m}^{\,\mathrm{content}}) \in \mathbb R^{C}$ while learning each $i$-th style word vector  $\mathbf{s}_{i}$.
Specifically, each style-content feature synthesized via $i$-th style word vector $\mathbf{s}_{i}$ should have the highest cosine similarity score with its corresponding content feature.
For $i$-th style word vector $\mathbf{s}_{i}$, a cosine similarity score $z_{imn}$ between a style-content feature with $m$-th class name and a content feature with $n$-th class name is computed by
\begin{align}
    \label{eq:logit_content}
    z_{imn} &= \frac{T(\mathcal{P}_{i}^{\,\mathrm{style}} \circ \mathcal{P}_{m}^{\,\mathrm{content}})}{\Vert T(\mathcal{P}_{i}^{\,\mathrm{style}} \circ \mathcal{P}_{m}^{\,\mathrm{content}}) \Vert_{\scriptscriptstyle{2}}} \bigcdot \frac{T(\mathcal{P}_{n}^{\,\mathrm{content}})}{\Vert T(\mathcal{P}_{n}^{\,\mathrm{content}}) \Vert_{\scriptscriptstyle{2}}} \;.
\end{align}
Using cosine similarity scores between style-content features and content features, the content consistency loss $\mathcal{L}_{\mathrm{content}}$ for learning $i$-th style word vector $\mathbf{s}_{i}$ is computed by 
\begin{align}
    \label{eq:loss_content}
    \mathcal{L}_{\mathrm{content}} &= -\frac{1}{N} \sum^{N}_{m=1} \log \left( \frac{\mathrm{exp}(z_{imm})}{\sum^{N}_{n=1}\mathrm{exp}(z_{imn})} \right) ,
\end{align}
where $N$ denotes the number of classes pre-defined in the target task.
This content loss $\mathcal{L}_{\mathrm{content}}$ is a contrastive loss which encourages each style-content feature to be located closer to its corresponding content feature so that it forces each $i$-th style word vector $\mathbf{s}_{i}$ to preserve content information when used to synthesize style-content features.

\noindent \textbf{Total prompt loss.}
\mbox{PromptStyler} learns $K$ style word vectors $\{\mathbf{s}_{i}\}^{K}_{i=1}$ in a sequential manner, where each $i$-th style word vector $\mathbf{s}_{i}$ is learned 
using both $\mathcal{L}_{\mathrm{style}}$ (Eq.\,(\ref{eq:loss_style})) and $\mathcal{L}_{\mathrm{content}}$ (Eq.\,(\ref{eq:loss_content})).
In the proposed method, the total loss $\mathcal{L}_{\mathrm{prompt}}$ for learning $i$-th style word vector is computed by
\begin{align}
    \label{eq:loss_prompt}
    \mathcal{L}_{\mathrm{prompt}} &= \mathcal{L}_{\mathrm{style}} + \mathcal{L}_{\mathrm{content}} \;.
\end{align}
Using this prompt loss $\mathcal{L}_{\mathrm{prompt}}$, we train $i$-th style word vector $\mathbf{s}_{i}$ for $L$ training iterations.

\subsection{Training a linear classifier using diverse styles}
After learning $K$ style word vectors $\{\mathbf{s}_{i}\}^{K}_{i=1}$, we generate $KN$ style-content features for training a linear classifier.
To be specific, we synthesize those features using the learned $K$ styles and pre-defined $N$ classes via the text encoder $T(\cdot)$.
The linear classifier is trained by a classification loss using $\ell_2$-normalized style-content features and their class labels;
each class label is the class name used to generate each style-content feature.
To effectively leverage the hyperspherical joint vision-language space,
we adopt ArcFace~\cite{ArcFace} loss as our classification loss $\mathcal{L}_{\mathrm{class}}$.
Note that ArcFace loss is an angular Softmax loss which computes the cosine similarities between classifier input features and classifier weights with an additive angular margin penalty between classes.
This angular margin penalty allows for more discriminative predictions by pushing features from different classes further apart.
Thanks to the property, this angular Softmax loss has been widely used in visual recognition tasks~\cite{ArcFace_app1,ArcFace_app2,ArcFace_app3,ArcFace_app4,ArcFace_app5}.

\begin{algorithm}[t!]
    \caption{\mbox{PromptStyler}}
    \label{alg:pseudocode_prompt}
    \textbf{Requirement:} pre-trained text encoder $T(\cdot)$, pre-defined $N$ \hspace*{\algorithmicindent}\hspace*{\algorithmicindent}\;\;\;\;\;\;\;\;\;\;\,\,class names in the target task \\
    \textbf{Input:} number of style word vectors $K$, number of training \hspace*{\algorithmicindent}\hspace*{\algorithmicindent}iterations $L$ \\
    \textbf{Output:}
    $KN$ style-content features
    \begin{algorithmic}[1]
        \Statex \commentcolor{\# randomly initialize style word vectors}
        
        \State 
        $\{\mathbf{s}_{i}\}^{K}_{i=1} \leftarrow \mathtt{random\_initialize}(\{\mathbf{s}_{i}\}^{K}_{i=1})$
        
        \Statex \commentcolor{\# sequentially learn $K$ style word vectors}
        
        \For {$i=1,2,\ldots,K$} 
        
            \Statex \commentcolor{\;\;\;\;\;\,\# $L$ training iterations for learning each word vector}
            
            \For {$\mathrm{iteration}=1,2,\ldots,L$} 
                
                \Statex \commentcolor{\;\;\;\;\;\;\;\;\;\;\,\,\# compute $\mathcal{L}_{\mathrm{style}}$ using $T(\cdot)$ and word vectors}
                
                \State $\mathcal{L}_{\mathrm{style}} \leftarrow \mathtt{style\_diversity\_loss}(\mathbf{s}_{i}, \{\mathbf{s}_{j}\}^{i-1}_{j=1})$
                
                \Statex \commentcolor{\;\;\;\;\;\;\;\;\;\;\,\,\# compute $\mathcal{L}_{\mathrm{content}}$ using $T(\cdot)$ and a word vector}
                
                \State $\mathcal{L}_{\mathrm{content}} \leftarrow \mathtt{content\_consistency\_loss}(\mathbf{s}_{i})$
                
                \State
                $\mathcal{L}_{\mathrm{prompt}} \leftarrow
                \mathcal{L}_{\mathrm{style}} + \mathcal{L}_{\mathrm{content}}$ 
                
                \State
                Update $\mathbf{s}_{i}$ using $\mathcal{L}_{\mathrm{prompt}}$ by gradient descent
                
            \EndFor
            
        \EndFor
        \State
        Synthesize $KN$ style-content features using the learned $K$ style word vectors and $N$ class names via $T(\cdot)$ 
    \end{algorithmic} 
\end{algorithm}

\subsection{Inference using the trained classifier}
The trained classifier is used with a pre-trained image encoder $I(\cdot)$ at inference time.
Given an input image $\mathbf{x}$, the image encoder extracts its image feature $I(\mathbf{x}) \in \mathbb{R}^{C}$, which is mapped to the hyperspherical joint vision-language space by $\ell_2$ normalization.
Then, the trained classifier produces class scores using the $\ell_2$-normalized image feature.
Note that the text encoder $T(\cdot)$ is not used at inference time, while the image encoder $I(\cdot)$ is only exploited at inference time.
\section{Experiments}
For more comprehensive understanding, please refer to the supplementary material (Section \redcolornumber{B} and \redcolornumber{D}).

\subsection{Evaluation datasets}
The proposed method does not require any actual data for training.
To analyze its generalization capability,
four domain generalization benchmarks are used for evaluation:
\textbf{PACS}~\cite{PACSdataset} (4 domains and 7 classes), \textbf{VLCS}~\cite{VLCSdataset} (4 domains and 5 classes), \textbf{OfficeHome}~\cite{OfficeHomedataset} (4 domains and 65 classes) and \textbf{DomainNet}~\cite{DomainNetdataset} (6 domains and 345 classes).
On these benchmarks, we repeat each experiment three times using different random seeds and report average top-1 classification accuracies with standard errors.
Unlike the \textit{leave-one-domain-out cross-validation} evaluation protocol~\cite{gulrajani2021domainbed}, we do not exploit any source domain data for training.

\subsection{Implementation details}
PromptStyler is implemented and trained with the same configuration regardless of the evaluation datasets.
Training takes about $30$ minutes using a single RTX 3090 GPU.

\noindent \textbf{Architecture.}
We choose CLIP~\cite{radford2021clip} as our large-scale pre-trained vision-language model, and use the publicly available pre-trained model.\footnote{\url{https://github.com/openai/CLIP}}
The text encoder $T(\cdot)$ used in training is Transformer~\cite{vaswani2017attention} and the image encoder $I(\cdot)$ used at inference is ResNet-50~\cite{resnet} as default setting in experiments; 
our method is also implemented with ViT-B/16~\cite{dosovitskiy2021an} or ViT-L/14~\cite{dosovitskiy2021an} for further evaluations as shown in Table~\ref{table:main_result}.
Note that text and image encoders are derived from the same CLIP model and frozen in the entire pipeline.
The dimension of each text feature or image feature is $C=1024$ when our method is implemented with ResNet-50, while $C=512$ in the case of ViT-B/16 and $C=768$ in the case of ViT-L/14.

\noindent \textbf{Learning style word vectors.}
We follow prompt learning methods~\cite{CoOp,CoCoOp} when learning the word vectors.
Using a zero-mean Gaussian distribution with $0.02$ standard deviation, 
we randomly initialize $K$ style word vectors $\{\mathbf{s}_{i}\}^{K}_{i=1}$, where $K=80$. 
The dimension of each style word vector is $D=512$ when the proposed method is implemented with ResNet-50~\cite{resnet} or ViT-B/16~\cite{dosovitskiy2021an}, while $D=768$ in the case of ViT-L/14~\cite{dosovitskiy2021an}.
Each $i$-th style word vector $\mathbf{s}_{i}$ is trained by the prompt loss $\mathcal{L}_{\mathrm{prompt}}$ for $L=100$ training iterations using the SGD optimizer with $0.002$ learning rate and $0.9$ momentum.
The number of classes $N$ is pre-defined by each target task definition, \eg, $N=345$ for DomainNet~\cite{DomainNetdataset}.

\noindent \textbf{Training a linear classifier.}
The classifier is trained for $50$ epochs using the SGD optimizer with $0.005$ learning rate, $0.9$ momentum, and a batch size of $128$.
In ArcFace~\cite{ArcFace} loss, its scaling factor is set to $5$ with $0.5$ angular margin.

\noindent \textbf{Inference.}
Input images are pre-processed in the same way with the CLIP model; resized to $224\times224$ and normalized.

\begin{table*}[!t]
    \centering
    \resizebox{\textwidth}{!}{
        \begin{tabular}{lccccccc|c}
        \Xhline{2\arrayrulewidth}
        \multicolumn{1}{c}{}
        & \multicolumn{2}{c}{Configuration}
        & \multicolumn{1}{c}{}
        & \multicolumn{5}{c}{Accuracy (\%)}
        \\
        \cline{2-3}
        \cline{5-9}

        \vspace{-0.8mm}
        & Source & Domain & \;
        & & & &
        &
        \\
        Method 
        & Domain & Description &
        & \normalsize{P}\small{ACS} & \normalsize{V}\small{LCS} & \normalsize{O}\small{fficeHome} & \normalsize{D}\small{omainNet} 
        & \normalsize{A}\small{vg.}
        \\
        \hline
        \multicolumn{9}{c}{\textit{ResNet-50~\cite{resnet} with pre-trained weights on ImageNet~\cite{deng2009imagenet}}}
        \\
        \hline
            DANN~\cite{DANN}
            & \ding{51} & \textbf{--} &
            & 83.6\scriptsize{$\pm{0.4}$} & 78.6\scriptsize{$\pm{0.4}$} & 65.9\scriptsize{$\pm{0.6}$} & 38.3\scriptsize{$\pm{0.1}$} 
            & 66.6
        \\
            RSC~\cite{RSC}
            & \ding{51} & \textbf{--} &
            & 85.2\scriptsize{$\pm{0.9}$} & 77.1\scriptsize{$\pm{0.5}$} & 65.5\scriptsize{$\pm{0.9}$} & 38.9\scriptsize{$\pm{0.5}$} 
            & 66.7
        \\
            MLDG~\cite{Li2018Learning}
            & \ding{51} & \textbf{--} &
            & 84.9\scriptsize{$\pm{1.0}$} & 77.2\scriptsize{$\pm{0.4}$} & 66.8\scriptsize{$\pm{0.6}$} & 41.2\scriptsize{$\pm{0.1}$} 
            & 67.5
        \\
            SagNet~\cite{SagNet}
            & \ding{51} & \textbf{--} &
            & \textbf{86.3}\scriptsize{$\pm{0.2}$} & 77.8\scriptsize{$\pm{0.5}$} & 68.1\scriptsize{$\pm{0.1}$} & 40.3\scriptsize{$\pm{0.1}$} 
            & 68.1
        \\
            SelfReg~\cite{SelfReg}
            & \ding{51} & \textbf{--} &
            & 85.6\scriptsize{$\pm{0.4}$} & 77.8\scriptsize{$\pm{0.9}$} & 67.9\scriptsize{$\pm{0.7}$} & 42.8\scriptsize{$\pm{0.0}$} 
            & 68.5
        \\
            GVRT~\cite{Min2022Grounding}
            & \ding{51} & \textbf{--} &
            & 85.1\scriptsize{$\pm{0.3}$} & \textbf{79.0}\scriptsize{$\pm{0.2}$} & 70.1\scriptsize{$\pm{0.1}$} & 44.1\scriptsize{$\pm{0.1}$} 
            & 69.6
        \\
            MIRO~\cite{cha2022miro}
            & \ding{51} & \textbf{--} &
            & 85.4\scriptsize{$\pm{0.4}$} & \textbf{79.0}\scriptsize{$\pm{0.0}$} & \textbf{70.5}\scriptsize{$\pm{0.4}$} & \textbf{44.3}\scriptsize{$\pm{0.2}$}
            & \textbf{69.8}
        \\
        \hhline{-|-|-|-|-|-|-|-|-|}
        \multicolumn{9}{c}{\textit{ResNet-50~\cite{resnet} with pre-trained weights from CLIP~\cite{radford2021clip}}}
        \\
        \hhline{-|-|-|-|-|-|-|-|-|}
            ZS-CLIP (C)~\cite{radford2021clip}\;\;\;\;
            & \textbf{--} & \textbf{--} &
            & 90.6\scriptsize{$\pm{0.0}$} & 76.0\scriptsize{$\pm{0.0}$} & 68.6\scriptsize{$\pm{0.0}$} & 45.6\scriptsize{$\pm{0.0}$} 
            & 70.2
        \\
            CAD~\cite{CAD}
            & \ding{51} & \textbf{--} &
            & 90.0\scriptsize{$\pm{0.6}$} & 81.2\scriptsize{$\pm{0.6}$} & 70.5\scriptsize{$\pm{0.3}$} & 45.5\scriptsize{$\pm{2.1}$} 
            & 71.8
        \\
            ZS-CLIP (PC)~\cite{radford2021clip}
            & \textbf{--} & \ding{51} &
            & 90.7\scriptsize{$\pm{0.0}$} & 80.1\scriptsize{$\pm{0.0}$} & 72.0\scriptsize{$\pm{0.0}$} & 46.2\scriptsize{$\pm{0.0}$} 
            & 72.3
        \\
            \cellcolor{gray!9.0}\textbf{PromptStyler}
            & \cellcolor{gray!9.0}\textbf{--} & \cellcolor{gray!9.0}\textbf{--} & \cellcolor{gray!9.0} & \cellcolor{gray!9.0}\textbf{93.2}\scriptsize{$\pm{0.0}$} & \cellcolor{gray!9.0}\textbf{82.3}\scriptsize{$\pm{0.1}$} & \cellcolor{gray!9.0}\textbf{73.6}\scriptsize{$\pm{0.1}$} & \cellcolor{gray!9.0}\textbf{49.5}\scriptsize{$\pm{0.0}$} 
            & \cellcolor{gray!9.0}\textbf{74.7}
        \\
        \hhline{-|-|-|-|-|-|-|-|-|}
        \multicolumn{9}{c}{\textit{ViT-B\,/\,16~\cite{dosovitskiy2021an} with pre-trained weights from CLIP~\cite{radford2021clip}}}
        \\
        \hhline{-|-|-|-|-|-|-|-|-|}
            ZS-CLIP (C)~\cite{radford2021clip}
            & \textbf{--} & \textbf{--} &
            & 95.7\scriptsize{$\pm{0.0}$} & 76.4\scriptsize{$\pm{0.0}$} & 79.9\scriptsize{$\pm{0.0}$} & 57.8\scriptsize{$\pm{0.0}$} 
            & 77.5
        \\
            MIRO~\cite{cha2022miro}
            & \ding{51} & \textbf{--} &
            & 95.6 & 82.2 & 82.5 & 54.0 
            & 78.6
        \\
            ZS-CLIP (PC)~\cite{radford2021clip}
            & \textbf{--} & \ding{51} &
            & 96.1\scriptsize{$\pm{0.0}$} & 82.4\scriptsize{$\pm{0.0}$} & 82.3\scriptsize{$\pm{0.0}$} & 57.7\scriptsize{$\pm{0.0}$} 
            & 79.6
        \\
            \cellcolor{gray!9.0}\textbf{PromptStyler}
            & \cellcolor{gray!9.0}\textbf{--} & \cellcolor{gray!9.0}\textbf{--} & \cellcolor{gray!9.0} & \cellcolor{gray!9.0}\textbf{97.2}\scriptsize{$\pm{0.1}$} & \cellcolor{gray!9.0}\textbf{82.9}\scriptsize{$\pm{0.0}$} & \cellcolor{gray!9.0}\textbf{83.6}\scriptsize{$\pm{0.0}$} & \cellcolor{gray!9.0}\textbf{59.4}\scriptsize{$\pm{0.0}$} 
            & \cellcolor{gray!9.0}\textbf{80.8}
        \\
        \hhline{-|-|-|-|-|-|-|-|-|}
        \multicolumn{9}{c}{\textit{ViT-L\,/\,14~\cite{dosovitskiy2021an} with pre-trained weights from CLIP~\cite{radford2021clip}}}
        \\
        \hhline{-|-|-|-|-|-|-|-|-|}
            ZS-CLIP (C)~\cite{radford2021clip}
            & \textbf{--} & \textbf{--} &
            & 97.6\scriptsize{$\pm{0.0}$} & 77.5\scriptsize{$\pm{0.0}$} & 85.9\scriptsize{$\pm{0.0}$} & 63.3\scriptsize{$\pm{0.0}$} 
            & 81.1
        \\
            ZS-CLIP (PC)~\cite{radford2021clip}
            & \textbf{--} & \ding{51} &
            & 98.5\scriptsize{$\pm{0.0}$} & \textbf{82.4}\scriptsize{$\pm{0.0}$} & 86.9\scriptsize{$\pm{0.0}$} & 64.0\scriptsize{$\pm{0.0}$} 
            & 83.0
        \\            
            \cellcolor{gray!9.0}\textbf{PromptStyler}
            & \cellcolor{gray!9.0}\textbf{--} & \cellcolor{gray!9.0}\textbf{--} & \cellcolor{gray!9.0} & \cellcolor{gray!9.0}\textbf{98.6}\scriptsize{$\pm{0.0}$} & \cellcolor{gray!9.0}\textbf{82.4}\scriptsize{$\pm{0.2}$} & \cellcolor{gray!9.0}\textbf{89.1}\scriptsize{$\pm{0.0}$} & \cellcolor{gray!9.0}\textbf{65.5}\scriptsize{$\pm{0.0}$} 
            & \cellcolor{gray!9.0}\textbf{83.9}
        \\
        \Xhline{2\arrayrulewidth}
    \end{tabular}}
    \vspace{-2mm}
    \caption{Comparison with the state-of-the-art domain generalization methods.
    ZS-CLIP (C) denotes zero-shot CLIP using ``[class]" as its text prompt, and ZS-CLIP (PC) indicates zero-shot CLIP using ``a photo of a [class]" as its text prompt.
    Note that \mbox{PromptStyler} does not exploit any source domain data and domain descriptions.}
    \vspace{0.25mm}
    \label{table:main_result}
\end{table*}

\subsection{Evaluations}

\noindent \textbf{Main results.}
PromptStyler achieves the state of the art in every evaluation on PACS~\cite{PACSdataset}, VLCS~\cite{VLCSdataset}, OfficeHome~\cite{OfficeHomedataset} and DomainNet~\cite{DomainNetdataset} as shown in Table~\ref{table:main_result}.
Note that all existing methods utilize source domain data except for zero-shot CLIP~\cite{radford2021clip} in Table~\ref{table:main_result}.
Compared with zero-shot CLIP which generates each text feature using a domain-agnostic prompt (``[class]"),
PromptStyler largely outperforms its records in all evaluations.
Our method also shows higher accuracy compared with zero-shot CLIP which produces each text feature using a domain-specific prompt (``a photo of a [class]"), 
even though we do not exploit any domain descriptions.
These results confirm that the proposed method effectively improves the generalization capability of the chosen pre-trained model, \ie, CLIP, without using any images by simulating various distribution shifts via prompts in its latent space.

\begin{table}[!t]
    \centering
    \resizebox{\columnwidth}{!}{
        \begin{tabular}{lcccc}
        \Xhline{2\arrayrulewidth}
        \multicolumn{1}{c}{}
        & \multicolumn{2}{c}{Inference Module}
        & \multicolumn{2}{c}{}
        \\
        \cline{2-3}
        \vspace{-0.5mm}
        & Image & Text & &
        \\
        Method 
        & Encoder & Encoder & \!\#\,Params\! & \!FPS\!
        \\
        \hline
        \multicolumn{5}{c}{\textit{\normalsize{O}\small{fficeHome} \normalsize{(65 classes)}}}
        \\
            \hhline{-|-|-|-|-|}
            ZS-CLIP~\cite{radford2021clip}
            & \ding{51} & \ding{51}
            & 102.0M
            & 1.6
        \\
            \cellcolor{gray!9.0}\textbf{PromptStyler}
            & \cellcolor{gray!9.0}\ding{51} & \cellcolor{gray!9.0}\textbf{--} 
            & \cellcolor{gray!9.0}\textbf{38.4M}
            & \cellcolor{gray!9.0}\textbf{72.9}
        \\
        \hhline{-|-|-|-|-|}
        \multicolumn{5}{c}{\textit{\normalsize{D}\small{omainNet} \normalsize{(345 classes)}}}
        \\
            \hhline{-|-|-|-|-|}
            ZS-CLIP~\cite{radford2021clip}
            & \ding{51} & \ding{51}
            & 102.0M
            & 0.3
        \\
            \cellcolor{gray!9.0}\textbf{PromptStyler}
            & \cellcolor{gray!9.0}\ding{51} & \cellcolor{gray!9.0}\textbf{--} 
            & \cellcolor{gray!9.0}\textbf{38.7M}
            & \cellcolor{gray!9.0}\textbf{72.9}
        \\
        \Xhline{2\arrayrulewidth}
    \end{tabular}}
    \vspace{-2mm}
    \caption{The number of parameters and inference speed on OfficeHome~\cite{OfficeHomedataset} and DomainNet~\cite{DomainNetdataset} using ResNet-50~\cite{resnet} as an image encoder. 
    Note that CLIP~\cite{radford2021clip} text encoder needs to generate text features as many as the number of classes.}
    \vspace{-5.5mm}
    \label{table:param_speed}
\end{table}

\begin{figure*}[t!]
    \centering
    \includegraphics[width=\textwidth]{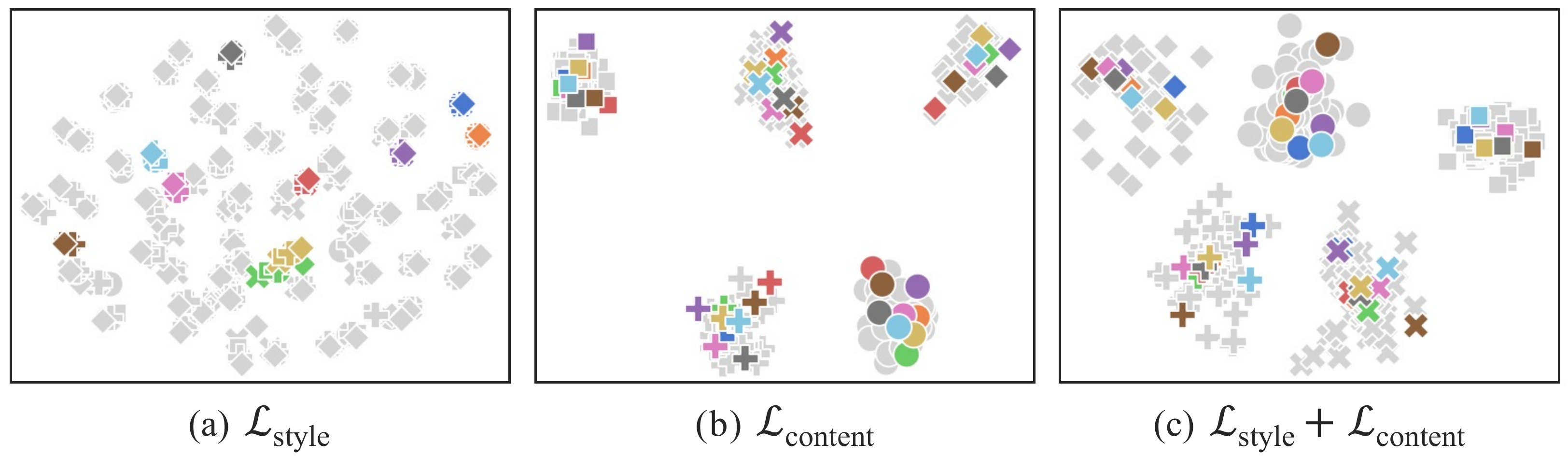}
    \vspace{-7mm}
    \caption{t-SNE~\cite{tSNE} visualization results for the target task VLCS~\cite{VLCSdataset} (5 classes) using synthesized style-content features.
    We visualize such features obtained from the learned $80$ style word vectors $\{\mathbf{s}_{i}\}^{80}_{i=1}$ and all the 5 classes (bird, car, chair, dog, person).
    Different colors denote features obtained from different style word vectors, and different shapes indicate features obtained from different class names.
    We only colorize features from the first $10$ styles $\{\mathbf{s}_{i}\}^{10}_{i=1}$.
    Combining the style diversity loss $\mathcal{L}_{\mathrm{style}}$ and content consistency loss $\mathcal{L}_{\mathrm{content}}$ leads to diverse styles while preserving content information.}
    \vspace{-1mm}
    \label{fig:fig4}
\end{figure*}

\begin{figure*}[t!]
    \centering
    \includegraphics[width=\textwidth]{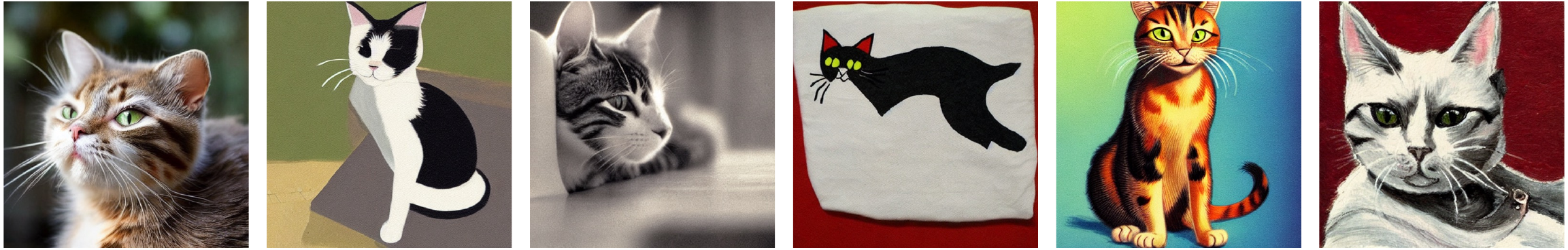}
    \vspace{-6.7mm}
    \caption{Text-to-Image synthesis results using style-content features (from ``a $\boldsymbol{S_{*}}$ style of a \textbf{cat}") with $6$ different style word vectors.
    By leveraging the proposed method,
    we could learn a variety of styles while not distorting content information.}
    \vspace{-3.1mm}
    \label{fig:fig5}
\end{figure*}

\noindent \textbf{Computational evaluations.}
In Table~\ref{table:param_speed}, we compare our PromptStyler and zero-shot CLIP~\cite{radford2021clip} in terms of the number of parameters and inference speed;
the inference speed was measured using a single RTX 3090 GPU with a batch size of $1$.
Note that we do not exploit a text encoder at inference time, 
which makes our model $\sim$2.6$\times$ smaller and $\sim$243$\times$ faster compared with CLIP.
Regarding the inference speed, the proposed model is about $45\times$ faster for the target task OfficeHome~\cite{OfficeHomedataset} ($65$ classes) and it is about $243\times$ faster for the target task DomainNet~\cite{DomainNetdataset} ($345$ classes).

\begin{table}[!t]
    \centering
    \resizebox{\columnwidth}{!}{
        \begin{tabular}{cccccc|c}
        \Xhline{2\arrayrulewidth}
        \multicolumn{2}{c}{}
        & \multicolumn{5}{c}{\small{Accuracy (\%)}}
        \\
        \cline{3-7}
        \!$\mathcal{L}_{\mathrm{style}}$\! & \!$\mathcal{L}_{\mathrm{content}}$\! 
        & \!\small{P}\scriptsize{ACS}\! & \!\small{V}\scriptsize{LCS}\! & \!\small{O}\scriptsize{fficeHome}\!\! & \!\small{D}\scriptsize{omainNet}\! & \!\small{A}\scriptsize{vg.}\!
        \\
        \hline
            \!\textbf{--}\! & \!\textbf{--}\! 
            & \!92.6\! & \!78.3\! & \!72.2\!\! & \!48.0\! & \!72.8\!
        \\
            \!\ding{51}\! & \!\textbf{--}\! 
            & \!92.3\! & \!80.9\! & \!71.5\!\! & \!48.2\! & \!73.2\!
        \\
            \!\textbf{--}\! & \!\ding{51}\!
            & \!92.8\! & \!80.5\! & \!72.4\!\! & \!48.6\! & \!73.6\!
        \\
            \!\cellcolor{gray!9.0}\ding{51}\! & \!\cellcolor{gray!9.0}\ding{51}\!
            & \cellcolor{gray!9.0}\!\textbf{93.2}\! & \cellcolor{gray!9.0}\!\textbf{82.3}\! & \cellcolor{gray!9.0}\!\textbf{73.6}\!\! & \cellcolor{gray!9.0}\!\textbf{49.5}\! & \cellcolor{gray!9.0}\!\textbf{74.7}\!
        \\
        \Xhline{2\arrayrulewidth}
    \end{tabular}}
    \vspace{-2mm}
    \caption{Ablation study on the style diversity loss $\mathcal{L}_{\mathrm{style}}$ and content consistency loss $\mathcal{L}_{\mathrm{content}}$ used in the prompt loss.}
    \vspace{-2.5mm}
    \label{table:ablation_loss}
\end{table}

\noindent \textbf{t-SNE visualization results.}
In Figure~\ref{fig:fig4}, we qualitatively evaluate style-content features synthesized for the target task VLCS~\cite{VLCSdataset} (5 classes) using t-SNE~\cite{tSNE} visualization.
As shown in Figure~\ref{fig:fig4}(c), PromptStyler generates a variety of styles while not distorting content information;
style-content features obtained from the same class name share similar semantics with diverse variations.
This result confirms that we could effectively simulate various distribution shifts in the latent space of a large-scale vision-language model by synthesizing diverse styles via learnable style word vectors.

\noindent \textbf{Text-to-Image synthesis results.}
In Figure~\ref{fig:fig5}, we visualize style-content features (from ``a $\boldsymbol{S_{*}}$ style of a \textbf{cat}") via diffusers library.\footnote{\url{https://github.com/huggingface/diffusers}} 
These results are obtained with $6$ different style word vectors, 
where the word vectors are learned for the target task DomainNet~\cite{DomainNetdataset} using ViT-L/14~\cite{dosovitskiy2021an} model.

\subsection{More analyses}
\vspace{-1mm}

\noindent \textbf{Ablation study on the prompt loss.}
In Table~\ref{table:ablation_loss}, we evaluate the effects of $\mathcal{L}_{\mathrm{style}}$ and $\mathcal{L}_{\mathrm{content}}$ in $\mathcal{L}_{\mathrm{prompt}}$ used for learning style words.
Interestingly, our method also achieves state-of-the-art results even without using these losses, \ie, the proposed framework (Fig.~\ref{fig:fig3}) is substantially effective by itself.
Note that randomly initialized style word vectors are already diverse, and CLIP~\cite{radford2021clip} is already good at extracting correct content information from a style-content prompt even without training the word vectors using $\mathcal{L}_{\mathrm{content}}$.
When we learn style word vectors using $\mathcal{L}_{\mathrm{style}}$ without  $\mathcal{L}_{\mathrm{content}}$, 
style-content features obtained from different class names share more similar features than those from the same class name (Fig.~\ref{fig:fig4}(a)).
On the other hand, using $\mathcal{L}_{\mathrm{content}}$ without $\mathcal{L}_{\mathrm{style}}$ leads to less diverse style-content features (Fig.~\ref{fig:fig4}(b)).
When incorporating both losses,
we could generate diverse styles while not distorting content information (Fig.~\ref{fig:fig4}(c)).

\begin{table}[!t]
    \centering
    \resizebox{\columnwidth}{!}{
        \begin{tabular}{ccccc|c}
        \Xhline{2\arrayrulewidth}
        \multicolumn{1}{c}{}
        & \multicolumn{5}{c}{\small{Accuracy (\%)}}
        \\
        \cline{2-6}
        $\mathcal{L}_{\mathrm{class}}$
        & \!\small{P}\scriptsize{ACS}\! & \!\small{V}\scriptsize{LCS}\! & \!\small{O}\scriptsize{fficeHome}\!\! & \!\small{D}\scriptsize{omainNet}\! & \!\small{A}\scriptsize{vg.}\!
        \\
        \hline
            \small{Softmax}
            & \!92.5\! & \!81.2\! & \!72.3\! & \!48.6\! & \!73.7\!
        \\
            \cellcolor{gray!9.0}\textbf{\small{ArcFace}}
            & \cellcolor{gray!9.0}\!\textbf{93.2}\! & \cellcolor{gray!9.0}\!\textbf{82.3}\! & \cellcolor{gray!9.0}\!\textbf{73.6}\! & \cellcolor{gray!9.0}\!\textbf{49.5}\! & \cellcolor{gray!9.0}\!\textbf{74.7}\!
        \\
        \Xhline{2\arrayrulewidth}
    \end{tabular}}
    \vspace{-2mm}
    \caption{Ablation study on the classification loss $\mathcal{L}_{\mathrm{class}}$ used for training a linear classifier in the proposed framework.}
    \vspace{-3.3mm}
    \label{table:ablation_classifier}
\end{table}

\begin{figure*}[t!]
    \centering
    \includegraphics[width=\textwidth]{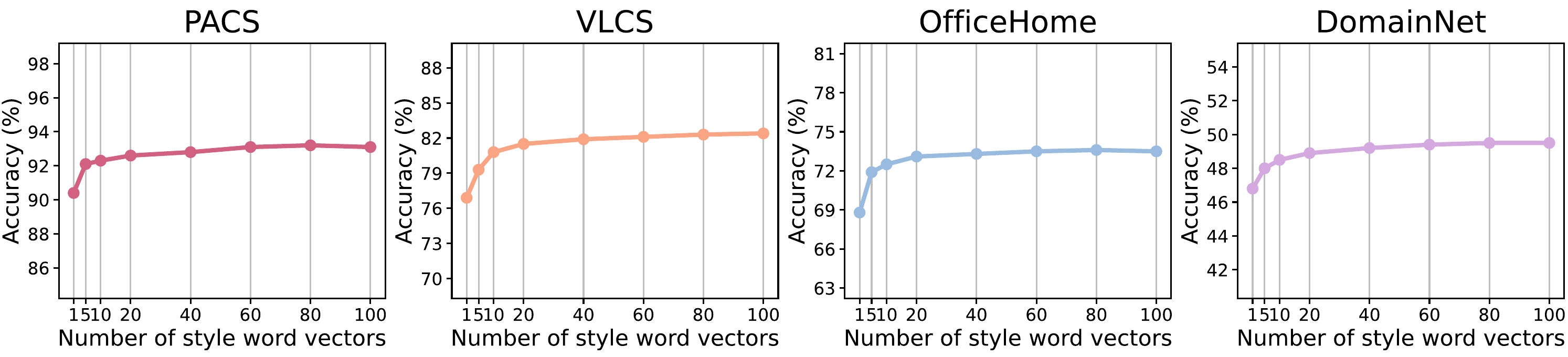}
    \vspace{-7.2mm}
    \caption{Top-1 classification accuracy on the PACS~\cite{PACSdataset}, VLCS~\cite{VLCSdataset}, OfficeHome~\cite{OfficeHomedataset} and DomainNet~\cite{DomainNetdataset} datasets with regard to the number of learnable style word vectors $K$.}
    \vspace{-3.5mm}
    \label{fig:fig6}
\end{figure*}

\begin{figure*}[t!]
    \centering
    \includegraphics[width=\textwidth]{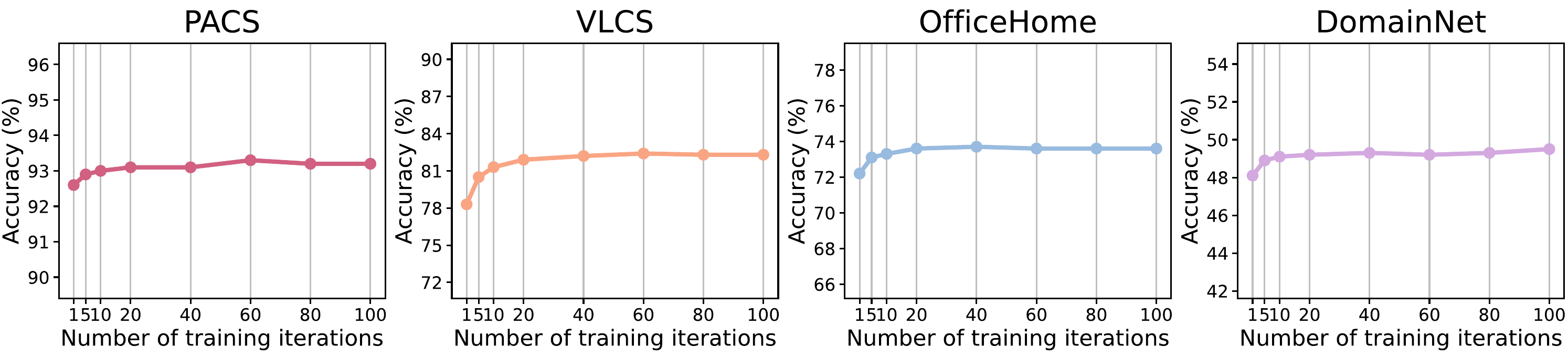}
    \vspace{-7.2mm}
    \caption{Top-1 classification accuracy on the PACS~\cite{PACSdataset}, VLCS~\cite{VLCSdataset}, OfficeHome~\cite{OfficeHomedataset} and DomainNet~\cite{DomainNetdataset} datasets with regard to the number of training iterations $L$ for learning each style word vector $\mathbf{s}_{i}$.}
    \vspace{-3.5mm}
    \label{fig:fig7}
\end{figure*}

\begin{table}[!t]
    \centering
    \resizebox{\columnwidth}{!}{
        \begin{tabular}{lccccc}
        \Xhline{2\arrayrulewidth}
        
        & \multicolumn{2}{c}{Configuration}
        & \multicolumn{1}{c}{}
        & \multicolumn{2}{c}{\!\!Accuracy (\%)\!\!}
        \\
        \cline{2-3}
        \cline{5-6}
        \vspace{-0.5mm}
        & \!\!Source\!\! & \!\!Domain\!\! &
        & \multicolumn{2}{c}{}
        \\
        Method
        & \!\!Domain\!\! & \!\!Description\!\! &
        & \multicolumn{2}{c}{\!\!\!\normalsize{T}\small{erra Incognita}\!\!\!}
        \\
        \hline
        \multicolumn{6}{c}{\textit{ResNet-50~\cite{resnet} with pre-trained weights on ImageNet~\cite{deng2009imagenet}}}
        \\
        \hhline{-|-|-|-|-|-|}
            SelfReg~\cite{SelfReg}
            & \ding{51} & \textbf{--} &
            & \multicolumn{2}{c}{47.0\scriptsize{$\pm{0.3}$}}
        \\
            GVRT~\cite{Min2022Grounding}
            & \ding{51} & \textbf{--} &
            & \multicolumn{2}{c}{\textbf{48.0}\scriptsize{$\pm{0.2}$}}
        \\
        \hhline{-|-|-|-|-|-|}
        \multicolumn{6}{c}{\textit{ResNet-50~\cite{resnet} with pre-trained weights from CLIP~\cite{radford2021clip}}}
        \\
        \hhline{-|-|-|-|-|-|}
            ZS-CLIP (C)~\cite{radford2021clip}\!\!
            & \textbf{--} & \textbf{--} &
            & \multicolumn{2}{c}{19.5\scriptsize{$\pm{0.0}$}}
        \\
            ZS-CLIP (PC)~\cite{radford2021clip}\!\!
            & \textbf{--} & \ding{51} &
            & \multicolumn{2}{c}{23.8\scriptsize{$\pm{0.0}$}}
        \\
            \cellcolor{gray!9.0}\textbf{PromptStyler}\!\!
            & \cellcolor{gray!9.0}\textbf{--} 
            & \cellcolor{gray!9.0}\textbf{--} 
            & \cellcolor{gray!9.0} 
            & \multicolumn{2}{c}{\cellcolor{gray!9.0}\textbf{30.5}\scriptsize{$\pm{0.8}$}}
        \\
        \Xhline{2\arrayrulewidth}
    \end{tabular}}
    \vspace{-2.5mm}
    \caption{Unsatisfactory results obtained from CLIP~\cite{radford2021clip} without using source domain data from Terra Incognita~\cite{TerraDataset}.}
    \vspace{-4.5mm}
    \label{table:terra_limitation}
\end{table}

\noindent \textbf{Ablation study on the classification loss.}
In Table~\ref{table:ablation_classifier}, we evaluate the effects of the original Softmax loss and the angular Softmax loss (\ie, ArcFace~\cite{ArcFace}).
PromptStyler also achieves the state of the art using the original one, 
which validates that the performance improvement of our method mainly comes from the proposed framework (Fig.~\ref{fig:fig3}).
Note that the angular Softmax loss further improves its accuracy by leveraging the hyperspherical joint vision-language space.

\noindent \textbf{Effect of the number of styles.}
We evaluate our method with regard to the number of style word vectors $K$ as shown in Figure~\ref{fig:fig6}.
Interestingly, our PromptStyler outperforms CLIP~\cite{radford2021clip} using just $5$ styles. 
This evaluation shows that $20$ style word vectors are enough to achieve decent results.

\noindent \textbf{Effect of the number of iterations.}
We evaluate our method with regard to the number of training iterations $L$ for learning each style word vector as shown in Figure~\ref{fig:fig7}.
This evaluation shows that $20$ iterations are enough to achieve decent results.
\section{Limitation}
\label{main:main_5}
\vspace{-0.5mm}
The performance of our method depends on the quality of the joint vision-language space constructed by the chosen vision-language model.
For example, although PromptStyler largely outperforms its base model (\ie, CLIP~\cite{radford2021clip}) in all evaluations, 
our method shows lower accuracy on the Terra Incognita dataset~\cite{TerraDataset} compared with other methods which utilize several images from the dataset as shown in Table~\ref{table:terra_limitation}.
The main reason for this might be due to the low accuracy of CLIP on the dataset.
Nevertheless, given that our method consistently outperforms its base model in every evaluation, this limitation could be alleviated with the development of large-scale vision-language models.
\section{Conclusion}
\vspace{-0.5mm}
We have presented a novel method that synthesizes a variety of styles in a joint vision-language space via learnable style words without exploiting any images to deal with source-free domain generalization.
PromptStyler simulates various distribution shifts in the latent space of a large-scale pre-trained model, which could effectively improve its generalization capability.
The proposed method achieves state-of-the-art results without using any source domain data on multiple domain generalization benchmarks.
We hope that future work could apply our method to other tasks using different large-scale vision-language models.

\vspace{1.5mm}
{\small
\noindent \textbf{Acknowledgment.} 
This work was supported by the Agency for Defense Development grant funded by the Korean government. 
}

{\small
\bibliographystyle{ieee_fullname}
\bibliography{main_arXiv}

\begin{thebibliography}{10}\itemsep=-1pt

\bibitem{TerraDataset}
Sara Beery, Grant van Horn, and Pietro Perona.
\newblock {Recognition in Terra Incognita}.
\newblock In {\em Proceedings of the European Conference on Computer Vision
  (ECCV)}, 2018.

\bibitem{ben2006da}
Shai Ben-David, John Blitzer, Koby Crammer, and Fernando Pereira.
\newblock {Analysis of Representations for Domain Adaptation}.
\newblock In {\em Advances in Neural Information Processing Systems (NIPS)},
  2006.

\bibitem{Carlucci2019Domain}
Fabio~Maria Carlucci, Antonio D'Innocente, Silvia Bucci, Barbara Caputo, and
  Tatiana Tommasi.
\newblock {Domain Generalization by Solving Jigsaw Puzzles}.
\newblock In {\em Proceedings of the IEEE/CVF Conference on Computer Vision and
  Pattern Recognition (CVPR)}, 2019.

\bibitem{SWAD}
Junbum Cha, Sanghyuk Chun, Kyungjae Lee, Han-Cheol Cho, Seunghyun Park, Yunsung
  Lee, and Sungrae Park.
\newblock {SWAD: Domain Generalization by Seeking Flat Minima}.
\newblock In {\em Advances in Neural Information Processing Systems (NeurIPS)},
  2021.

\bibitem{cha2022miro}
Junbum Cha, Kyungjae Lee, Sungrae Park, and Sanghyuk Chun.
\newblock {Domain Generalization by Mutual-Information Regularization with
  Pre-trained Models}.
\newblock In {\em Proceedings of the European Conference on Computer Vision
  (ECCV)}, 2022.

\bibitem{deng2009imagenet}
Jia Deng, Wei Dong, Richard Socher, Li-Jia Li, Kai Li, and Li Fei-Fei.
\newblock {ImageNet: A large-scale hierarchical image database}.
\newblock In {\em 2009 IEEE Conference on Computer Vision and Pattern
  Recognition}, pages 248--255, 2009.

\bibitem{ArcFace_app3}
Jiankang Deng, Jia Guo, Tongliang Liu, Mingming Gong, and Stefanos Zafeiriou.
\newblock {Sub-center ArcFace: Boosting Face Recognition by Large-Scale Noisy
  Web Faces}.
\newblock In {\em Proceedings of the European Conference on Computer Vision
  (ECCV)}, 2020.

\bibitem{ArcFace}
Jiankang Deng, Jia Guo, Niannan Xue, and Stefanos Zafeiriou.
\newblock {ArcFace: Additive Angular Margin Loss for Deep Face Recognition}.
\newblock In {\em Proceedings of the IEEE/CVF Conference on Computer Vision and
  Pattern Recognition (CVPR)}, 2019.

\bibitem{ArcFace_app1}
Jiankang Deng and Stefanos Zafeririou.
\newblock {ArcFace for Disguised Face Recognition}.
\newblock In {\em Proceedings of the IEEE/CVF International Conference on
  Computer Vision (ICCV)}, 2019.

\bibitem{LRDG2022}
Yu Ding, Lei Wang, Bin Liang, Shuming Liang, Yang Wang, and Fang Chen.
\newblock {Domain Generalization by Learning and Removing Domain-specific
  Features}.
\newblock In {\em Advances in Neural Information Processing Systems (NeurIPS)},
  2022.

\bibitem{dosovitskiy2021an}
Alexey Dosovitskiy, Lucas Beyer, Alexander Kolesnikov, Dirk Weissenborn,
  Xiaohua Zhai, Thomas Unterthiner, Mostafa Dehghani, Matthias Minderer, Georg
  Heigold, Sylvain Gelly, Jakob Uszkoreit, and Neil Houlsby.
\newblock {An Image is Worth 16x16 Words: Transformers for Image Recognition at
  Scale}.
\newblock In {\em International Conference on Learning Representations (ICLR)},
  2021.

\bibitem{Dou2019Domain}
Qi Dou, Daniel~C. Castro, Konstantinos Kamnitsas, and Ben Glocker.
\newblock {Domain Generalization via Model-Agnostic Learning of Semantic
  Features}.
\newblock In {\em Advances in Neural Information Processing Systems (NeurIPS)},
  2019.

\bibitem{LADS}
Lisa Dunlap, Clara Mohri, Devin Guillory, Han Zhang, Trevor Darrell, Joseph~E.
  Gonzalez, Aditi Raghunathan, and Anja Rohrbach.
\newblock {Using Language to Extend to Unseen Domains}.
\newblock In {\em International Conference on Learning Representations (ICLR)},
  2023.

\bibitem{Fan2021Adversarially}
Xinjie Fan, Qifei Wang, Junjie Ke, Feng Yang, Boqing Gong, and Mingyuan Zhou.
\newblock {Adversarially Adaptive Normalization for Single Domain
  Generalization}.
\newblock In {\em Proceedings of the IEEE/CVF Conference on Computer Vision and
  Pattern Recognition (CVPR)}, 2021.

\bibitem{VLCSdataset}
Chen Fang, Ye Xu, and Daniel~N. Rockmore.
\newblock {Unbiased Metric Learning: On the Utilization of Multiple Datasets
  and Web Images for Softening Bias}.
\newblock In {\em Proceedings of the IEEE International Conference on Computer
  Vision (ICCV)}, 2013.

\bibitem{frikha2022data}
Ahmed Frikha, Haokun Chen, Denis Krompaß, Thomas Runkler, and Volker Tresp.
\newblock {Towards Data-Free Domain Generalization}.
\newblock In {\em Asian Conference on Machine Learning (ACML)}, 2022.

\bibitem{gal2023textualinversion}
Rinon Gal, Yuval Alaluf, Yuval Atzmon, Or Patashnik, Amit~H. Bermano, Gal
  Chechik, and Daniel Cohen-Or.
\newblock {An Image is Worth One Word: Personalizing Text-to-Image Generation
  using Textual Inversion}.
\newblock In {\em International Conference on Learning Representations (ICLR)},
  2023.

\bibitem{StyleGAN_NADA}
Rinon Gal, Or Patashnik, Haggai Maron, Gal Chechik, and Daniel Cohen-Or.
\newblock {StyleGAN-NADA: CLIP-Guided Domain Adaptation of Image Generators}.
\newblock {\em ACM Transactions on Graphics (Proc. SIGGRAPH Asia)}, 2022.

\bibitem{DANN}
Yaroslav Ganin, Evgeniya Ustinova, Hana Ajakan, Pascal Germain, Hugo
  Larochelle, François Laviolette, Mario Marchand, and Victor Lempitsky.
\newblock {Domain-Adversarial Training of Neural Networks}.
\newblock In {\em Journal of Machine Learning Research (JMLR)}, 2016.

\bibitem{CLIPAdapter}
Peng Gao, Shijie Geng, Renrui Zhang, Teli Ma, Rongyao Fang, Yongfeng Zhang,
  Hongsheng Li, and Yu Qiao.
\newblock {CLIP-Adapter: Better Vision-Language Models with Feature Adapters}.
\newblock {\em arXiv preprint arXiv:2110.04544}, 2021.

\bibitem{gulrajani2021domainbed}
Ishaan Gulrajani and David Lopez-Paz.
\newblock {In Search of Lost Domain Generalization}.
\newblock In {\em International Conference on Learning Representations (ICLR)},
  2021.

\bibitem{resnet}
{He, Kaiming and Zhang, Xiangyu and Ren, Shaoqing and Sun, Jian}.
\newblock {Deep Residual Learning for Image Recognition}.
\newblock In {\em Proceedings of the IEEE Conference on Computer Vision and
  Pattern Recognition (CVPR)}, pages 770--778, 2016.

\bibitem{hendrycks2019ood}
Dan Hendrycks and Thomas Dietterich.
\newblock {Benchmarking Neural Network Robustness to Common Corruptions and
  Perturbations}.
\newblock In {\em International Conference on Learning Representations (ICLR)},
  2019.

\bibitem{hoffman2018cycada}
Judy Hoffman, Eric Tzeng, Taesung Park, Jun-Yan Zhu, Phillip Isola, Kate
  Saenko, Alexei~A. Efros, and Trevor Darrell.
\newblock {CyCADA: Cycle-Consistent Adversarial Domain Adaptation}.
\newblock In {\em International Conference on Machine Learning (ICML)}, 2018.

\bibitem{RSC}
Zeyi Huang, Haohan Wang, Eric~P. Xing, and Dong Huang.
\newblock {Self-Challenging Improves Cross-Domain Generalization}.
\newblock In {\em Proceedings of the European Conference on Computer Vision
  (ECCV)}, 2020.

\bibitem{ALIGN}
Chao Jia, Yinfei Yang, Ye Xia, Yi-Ting Chen, Zarana Parekh, Hieu Pham, Quoc~V.
  Le, Yunhsuan Sung, Zhen Li, and Tom Duerig.
\newblock {Scaling Up Visual and Vision-Language Representation Learning With
  Noisy Text Supervision}.
\newblock In {\em International Conference on Machine Learning (ICML)}, 2021.

\bibitem{kang2022styleneophile}
Juwon Kang, Sohyun Lee, Namyup Kim, and Suha Kwak.
\newblock {Style Neophile: Constantly Seeking Novel Styles for Domain
  Generalization}.
\newblock In {\em Proceedings of the IEEE/CVF Conference on Computer Vision and
  Pattern Recognition (CVPR)}, 2022.

\bibitem{SelfReg}
Daehee Kim, Seunghyun Park, Jinkyu Kim, and Jaekoo Lee.
\newblock {SelfReg: Self-supervised Contrastive Regularization for Domain
  Generalization}.
\newblock In {\em Proceedings of the IEEE/CVF International Conference on
  Computer Vision (ICCV)}, 2021.

\bibitem{kim2022broad}
Donghyun Kim, Kaihong Wang, Stan Sclaroff, and Kate Saenko.
\newblock {A Broad Study of Pre-training for Domain Generalization and
  Adaptation}.
\newblock In {\em Proceedings of the European Conference on Computer Vision
  (ECCV)}, 2022.

\bibitem{ArcFace_app4}
Dimitrios Kollias and Stefanos Zafeiriou.
\newblock {Expression, Affect, Action Unit Recognition: Aff-Wild2, Multi-Task
  Learning and ArcFace}.
\newblock In {\em Proceedings of the British Machine Vision Conference (BMVC)},
  2019.

\bibitem{CLIPstyler}
Gihyun Kwon and Jong~Chul Ye.
\newblock {CLIPstyler: Image Style Transfer with a Single Text Condition}.
\newblock In {\em Proceedings of the IEEE/CVF Conference on Computer Vision and
  Pattern Recognition (CVPR)}, 2022.

\bibitem{lee2022fifo}
Sohyun Lee, Taeyoung Son, and Suha Kwak.
\newblock {FIFO: Learning Fog-invariant Features for Foggy Scene Segmentation}.
\newblock In {\em Proceedings of the IEEE/CVF Conference on Computer Vision and
  Pattern Recognition (CVPR)}, 2022.

\bibitem{lee2023surgical}
Yoonho Lee, Annie~S. Chen, Fahim Tajwar, Ananya Kumar, Huaxiu Yao, Percy Liang,
  and Chelsea Finn.
\newblock {Surgical Fine-Tuning Improves Adaptation to Distribution Shifts}.
\newblock In {\em International Conference on Learning Representations (ICLR)},
  2023.

\bibitem{PACSdataset}
Da Li, Yongxin Yang, Yi-Zhe Song, and Timothy~M. Hospedales.
\newblock {Deeper, Broader and Artier Domain Generalization}.
\newblock In {\em Proceedings of the IEEE International Conference on Computer
  Vision (ICCV)}, 2017.

\bibitem{Li2018Learning}
Da Li, Yongxin Yang, Yi-Zhe Song, and Timothy~M. Hospedales.
\newblock {Learning to Generalize: Meta-Learning for Domain Generalization}.
\newblock In {\em Proceedings of the AAAI Conference on Artificial Intelligence
  (AAAI)}, 2018.

\bibitem{Li2019Episodic}
Da Li, Jianshu Zhang, Yongxin Yang, Cong Liu, Yi-Zhe Song, and Timothy~M.
  Hospedales.
\newblock {Episodic Training for Domain Generalization}.
\newblock In {\em Proceedings of the IEEE/CVF International Conference on
  Computer Vision (ICCV)}, 2019.

\bibitem{Li2018Domain}
Haoliang Li, Sinno~Jialin Pan, Shiqi Wang, and Alex~C. Kot.
\newblock {Domain Generalization With Adversarial Feature Learning}.
\newblock In {\em Proceedings of the IEEE/CVF Conference on Computer Vision and
  Pattern Recognition (CVPR)}, 2018.

\bibitem{Li2021Progressive}
Lei Li, Ke Gao, Juan Cao, Ziyao Huang, Yepeng Weng, Xiaoyue Mi, Zhengze Yu,
  Xiaoya Li, and Boyang xia.
\newblock {Progressive Domain Expansion Network for Single Domain
  Generalization}.
\newblock In {\em Proceedings of the IEEE/CVF Conference on Computer Vision and
  Pattern Recognition (CVPR)}, 2021.

\bibitem{liang2022mind}
Weixin Liang, Yuhui Zhang, Yongchan Kwon, Serena Yeung, and James Zou.
\newblock {Mind the Gap: Understanding the Modality Gap in Multi-modal
  Contrastive Representation Learning}.
\newblock In {\em Advances in Neural Information Processing Systems (NeurIPS)},
  2022.

\bibitem{ArcFace_app2}
Boxiao Liu, Guanglu Song, Manyuan Zhang, Haihang You, and Yu Liu.
\newblock {Switchable K-Class Hyperplanes for Noise-Robust Representation
  Learning}.
\newblock In {\em Proceedings of the IEEE/CVF International Conference on
  Computer Vision (ICCV)}, 2021.

\bibitem{ProDA}
Yuning Lu, Jianzhuang Liu, Yonggang Zhang, Yajing Liu, and Xinmei Tian.
\newblock {Prompt Distribution Learning}.
\newblock In {\em Proceedings of the IEEE/CVF Conference on Computer Vision and
  Pattern Recognition (CVPR)}, 2022.

\bibitem{Mahajan2021Domain}
Divyat Mahajan, Shruti Tople, and Amit Sharma.
\newblock {Domain Generalization using Causal Matching}.
\newblock In {\em International Conference on Machine Learning (ICML)}, 2021.

\bibitem{Matsuura2020Domain}
Toshihiko Matsuura and Tatsuya Harada.
\newblock {Domain Generalization Using a Mixture of Multiple Latent Domains}.
\newblock In {\em Proceedings of the AAAI Conference on Artificial Intelligence
  (AAAI)}, 2020.

\bibitem{Min2022Grounding}
Seonwoo Min, Nokyung Park, Siwon Kim, Seunghyun Park, and Jinkyu Kim.
\newblock {Grounding Visual Representations with Texts for Domain
  Generalization}.
\newblock In {\em Proceedings of the European Conference on Computer Vision
  (ECCV)}, 2022.

\bibitem{Muandet2013DG}
Krikamol Muandet, David Balduzzi, and Bernhard Schölkopf.
\newblock {Domain Generalization via Invariant Feature Representation}.
\newblock In {\em International Conference on Machine Learning (ICML)}, 2013.

\bibitem{SagNet}
Hyeonseob Nam, HyunJae Lee, Jongchan Park, Wonjun Yoon, and Donggeun Yoo.
\newblock {Reducing Domain Gap by Reducing Style Bias}.
\newblock In {\em Proceedings of the IEEE/CVF Conference on Computer Vision and
  Pattern Recognition (CVPR)}, 2021.

\bibitem{StlyeCLIP}
Or Patashnik, Zongze Wu, Eli Shechtman, Daniel Cohen-Or, and Dani Lischinski.
\newblock {StyleCLIP: Text-Driven Manipulation of StyleGAN Imagery}.
\newblock In {\em Proceedings of the IEEE/CVF International Conference on
  Computer Vision (ICCV)}, 2021.

\bibitem{DomainNetdataset}
Xingchao Peng, Qinxun Bai, Xide Xia, Zijun Huang, Kate Saenko, and Bo Wang.
\newblock {Moment Matching for Multi-Source Domain Adaptation}.
\newblock In {\em Proceedings of the IEEE/CVF International Conference on
  Computer Vision (ICCV)}, 2019.

\bibitem{Qiao2020Learning}
Fengchun Qiao, Long Zhao, and Xi Peng.
\newblock {Learning to Learn Single Domain Generalization}.
\newblock In {\em Proceedings of the IEEE/CVF Conference on Computer Vision and
  Pattern Recognition (CVPR)}, 2020.

\bibitem{radford2021clip}
Alec Radford, Jong~Wook Kim, Chris Hallacy, Aditya Ramesh, Gabriel Goh,
  Sandhini Agarwal, Girish Sastry, Amanda Askell, Pamela Mishkin, Jack Clark,
  Gretchen Krueger, and Ilya Sutskever.
\newblock {Learning Transferable Visual Models From Natural Language
  Supervision}.
\newblock In {\em International Conference on Machine Learning (ICML)}, 2021.

\bibitem{Rame2022Fishr}
Alexandre Rame, Corentin Dancette, and Matthieu Cord.
\newblock {Fishr: Invariant Gradient Variances for Out-of-Distribution
  Generalization}.
\newblock In {\em International Conference on Machine Learning (ICML)}, 2022.

\bibitem{recht2019ood}
Benjamin Recht, Rebecca Roelofs, Ludwig Schmidt, and Vaishaal Shankar.
\newblock {Do ImageNet Classifiers Generalize to ImageNet?}
\newblock In {\em International Conference on Machine Learning (ICML)}, 2019.

\bibitem{CAD}
Yangjun Ruan, Yann Dubois, and Chris~J. Maddison.
\newblock {Optimal Representations for Covariate Shift}.
\newblock In {\em International Conference on Learning Representations (ICLR)},
  2022.

\bibitem{Saito2019semi}
Kuniaki Saito, Donghyun Kim, Stan Sclaroff, Trevor Darrell, and Kate Saenko.
\newblock {Semi-Supervised Domain Adaptation via Minimax Entropy}.
\newblock In {\em Proceedings of the IEEE/CVF International Conference on
  Computer Vision (ICCV)}, 2019.

\bibitem{Seo2020Learning}
Seonguk Seo, Yumin Suh, Dongwan Kim, Geeho Kim, Jongwoo Han, and Bohyung Han.
\newblock {Learning to Optimize Domain Specific Normalization for Domain
  Generalization}.
\newblock In {\em Proceedings of the European Conference on Computer Vision
  (ECCV)}, 2020.

\bibitem{sun2016coral}
Baochen Sun, Jiashi Feng, and Kate Saenko.
\newblock {Return of Frustratingly Easy Domain Adaptation}.
\newblock In {\em Proceedings of the AAAI Conference on Artificial Intelligence
  (AAAI)}, 2016.

\bibitem{tzeng2017adversarial}
Eric Tzeng, Judy Hoffman, Kate Saenko, and Trevor Darrell.
\newblock {Adversarial Discriminative Domain Adaptation}.
\newblock In {\em Proceedings of the IEEE Conference on Computer Vision and
  Pattern Recognition (CVPR)}, 2017.

\bibitem{tSNE}
Laurens van~der Maaten and Geoffrey Hinton.
\newblock {Visualizing Data using t-SNE}.
\newblock In {\em Journal of Machine Learning Research (JMLR)}, 2008.

\bibitem{vaswani2017attention}
Ashish Vaswani, Noam Shazeer, Niki Parmar, Jakob Uszkoreit, Llion Jones,
  Aidan~N Gomez, \L{}ukasz Kaiser, and Illia Polosukhin.
\newblock {Attention is All you Need}.
\newblock In {\em Advances in Neural Information Processing Systems (NIPS)},
  2017.

\bibitem{OfficeHomedataset}
Hemanth Venkateswara, Jose Eusebio, Shayok Chakraborty, and Sethuraman
  Panchanathan.
\newblock {Deep Hashing Network for Unsupervised Domain Adaptation}.
\newblock In {\em Proceedings of the IEEE Conference on Computer Vision and
  Pattern Recognition (CVPR)}, 2017.

\bibitem{wang2021dgsurvey}
Jindong Wang, Cuiling Lan, Chang Liu, Yidong Ouyang, Tao Qin, Wang Lu, Yiqiang
  Chen, Wenjun Zeng, and Philip~S. Yu.
\newblock {Generalizing to Unseen Domains: A Survey on Domain Generalization}.
\newblock In {\em IEEE Transactions on Knowledge and Data Engineering (TKDE)},
  2021.

\bibitem{Wang2021Learning}
Zijian Wang, Yadan Luo, Ruihong Qiu, Zi Huang, and Mahsa Baktashmotlagh.
\newblock {Learning to Diversify for Single Domain Generalization}.
\newblock In {\em Proceedings of the IEEE/CVF International Conference on
  Computer Vision (ICCV)}, 2021.

\bibitem{Xu2021FACT}
Qinwei Xu, Ruipeng Zhang, Ya Zhang, Yanfeng Wang, and Qi Tian.
\newblock {A Fourier-based Framework for Domain Generalization}.
\newblock In {\em Proceedings of the IEEE Conference on Computer Vision and
  Pattern Recognition (CVPR)}, 2021.

\bibitem{TCL}
Jinyu Yang, Jiali Duan, Son Tran, Yi Xu, Sampath Chanda, Liqun Chen, Belinda
  Zeng, Trishul Chilimbi, and Junzhou Huang.
\newblock {Vision-Language Pre-Training with Triple Contrastive Learning}.
\newblock In {\em Proceedings of the IEEE/CVF Conference on Computer Vision and
  Pattern Recognition (CVPR)}, 2022.

\bibitem{ArcFace_app5}
Dingyi Zhang, Yingming Li, and Zhongfei Zhang.
\newblock {Deep metric learning with spherical embedding}.
\newblock In {\em Advances in Neural Information Processing Systems (NeurIPS)},
  2020.

\bibitem{TipAdapter}
Renrui Zhang, Zhang Wei, Rongyao Fang, Peng Gao, Kunchang Li, Jifeng Dai, Yu
  Qiao, and Hongsheng Li.
\newblock {Tip-Adapter: Training-free Adaption of CLIP for Few-shot
  Classification}.
\newblock In {\em Proceedings of the European Conference on Computer Vision
  (ECCV)}, 2022.

\bibitem{zhang2023diagnosing}
Yuhui Zhang, Jeff~Z. HaoChen, Shih-Cheng Huang, Kuan-Chieh Wang, James Zou, and
  Serena Yeung.
\newblock {Diagnosing and Rectifying Vision Models using Language}.
\newblock In {\em International Conference on Learning Representations (ICLR)},
  2023.

\bibitem{zhao2019da}
Han Zhao, Remi Tachet~des Combes, Kun Zhang, and Geoffrey~J. Gordon.
\newblock {On Learning Invariant Representation for Domain Adaptation}.
\newblock In {\em International Conference on Machine Learning (ICML)}, 2019.

\bibitem{zhou2022dgsurvey}
Kaiyang Zhou, Ziwei Liu, Yu Qiao, Tao Xiang, and Chen~Change Loy.
\newblock {Domain Generalization: A Survey}.
\newblock In {\em IEEE Transactions on Pattern Analysis and Machine
  Intelligence (TPAMI)}, 2022.

\bibitem{CoCoOp}
Kaiyang Zhou, Jingkang Yang, Chen~Change Loy, and Ziwei Liu.
\newblock {Conditional Prompt Learning for Vision-Language Models}.
\newblock In {\em Proceedings of the IEEE/CVF Conference on Computer Vision and
  Pattern Recognition (CVPR)}, 2022.

\bibitem{CoOp}
Kaiyang Zhou, Jingkang Yang, Chen~Change Loy, and Ziwei Liu.
\newblock {Learning to Prompt for Vision-Language Models}.
\newblock In {\em International Journal of Computer Vision (IJCV)}, 2022.

\bibitem{zhou2020ddaig}
Kaiyang Zhou, Yongxin Yang, Timothy Hospedales, and Tao Xiang.
\newblock {Deep Domain-Adversarial Image Generation for Domain Generalisation}.
\newblock In {\em Proceedings of the AAAI Conference on Artificial Intelligence
  (AAAI)}, 2020.

\bibitem{Zhou2020Learning}
Kaiyang Zhou, Yongxin Yang, Timothy Hospedales, and Tao Xiang.
\newblock {Learning to Generate Novel Domains for Domain Generalization}.
\newblock In {\em Proceedings of the European Conference on Computer Vision
  (ECCV)}, 2020.

\bibitem{zhou2021mixstyle}
Kaiyang Zhou, Yongxin Yang, Yu Qiao, and Tao Xiang.
\newblock {Domain Generalization with MixStyle}.
\newblock In {\em International Conference on Learning Representations (ICLR)},
  2021.

\end{thebibliography}
}

\clearpage
\setcounter{section}{0}
\setcounter{figure}{0}
\setcounter{table}{0}
\renewcommand{\thesection}{\Alph{section}}

\title{PromptStyler: Prompt-driven Style Generation \\
for Source-free Domain Generalization \\
\textmd{--- Supplementary Material ---}}
\author{Junhyeong Cho$^1$ \quad Gilhyun Nam$^1$ \quad Sungyeon Kim$^2$ \quad Hunmin Yang$^{1,3}$ \quad Suha Kwak$^2$\\
{\small $^1$ADD \qquad $^2$POSTECH \qquad $^3$KAIST}\\
{\small \url{https://PromptStyler.github.io}}}
\date{}

\maketitle 
In this supplementary material, we provide more method details (Section~\redcolornumber{A}), analyses on Terra Incognita (Section~\redcolornumber{B}), evaluation results (Section~\redcolornumber{C}) and discussion (Section~\redcolornumber{D}).
\setcounter{figure}{0}
\setcounter{table}{0}
\renewcommand\thefigure{A\arabic{figure}}
\renewcommand\thetable{A\arabic{table}}

\section{Method Details}
\label{supp:supp_1}
This section provides more details of the chosen vision-language model (Section~\redcolornumber{A.1}) and design choices for learning style word vectors (Section~\redcolornumber{A.2}).

\subsection{Large-scale vision-language model}
\label{supp:subsection_vlm}

We choose CLIP~\cite{radford2021clip} as our pre-trained vision-language model which is a large-scale model trained with 400 million image-text pairs.
Note that the proposed method is broadly applicable to the CLIP-like vision-language models~\cite{ALIGN,TCL} which also construct hyperspherical joint vision-language spaces using contrastive learning methods. 
Given a batch of image-text pairs, such models jointly train an image encoder and a text encoder considering similarity scores obtained from image-text pairings.

\noindent \textbf{Joint vision-language training.}
Suppose there is a batch of $M$ image-text pairs.
Among all possible $M \times M$ pairings,
the matched $M$ pairs are the positive pairs and the other $M^{2}-M$ pairs are the negative pairs.
CLIP~\cite{radford2021clip} is trained to maximize cosine similarities of image and text features from the positive $M$ pairs while minimizing the similarities of such features from the negative $M^{2}-M$ pairs.

\noindent \textbf{Image encoder.}
CLIP~\cite{radford2021clip} utilizes ResNet~\cite{resnet} or ViT~\cite{dosovitskiy2021an} as its image encoder.
Given an input image, the image encoder extracts its image feature.
After that, the image feature is mapped to a hyperspherical joint vision-language space by $\ell_{2}$ normalization.

\noindent \textbf{Text encoder.}
CLIP~\cite{radford2021clip} utilizes Transformer~\cite{vaswani2017attention} as its text encoder.
Given an input text prompt, it is converted to word vectors via a tokenization process and a word lookup procedure.
Using these word vectors, the text encoder generates a text feature which is then mapped to a hyperspherical joint vision-language space by $\ell_{2}$ normalization.

\noindent \textbf{Zero-shot inference.}
At inference time, zero-shot CLIP~\cite{radford2021clip} synthesizes classifier weights via the text encoder using $N$ class names pre-defined in the target task.
Given an input image, the image encoder extracts its image feature and the text encoder produces $N$ text features using the $N$ class names.
Then, it computes cosine similarity scores between the image feature and text features, and selects the class name which results in the highest similarity score as its classification output.

\begin{figure}[t!]
    \centering
    \includegraphics[width=\columnwidth]{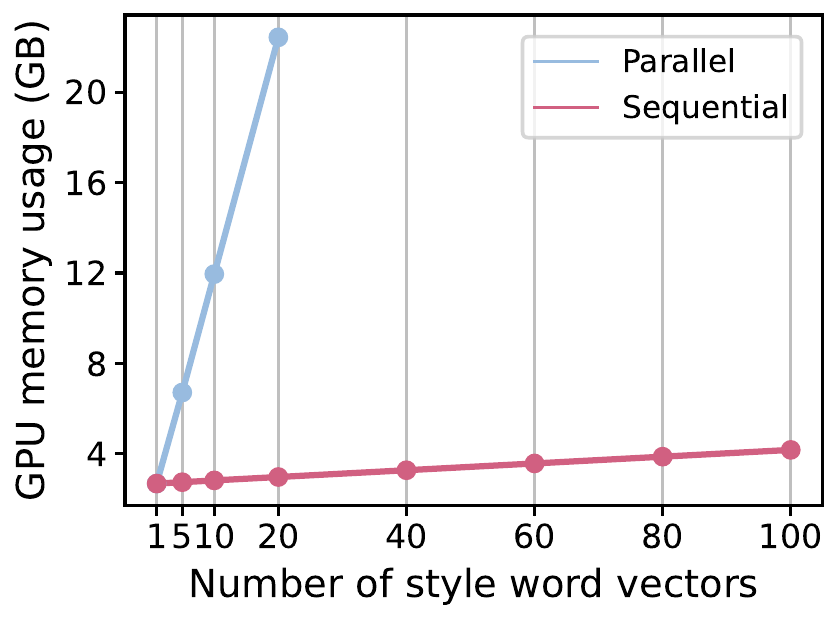}
    \vspace{-7.5mm}
    \caption{GPU memory usage when learning $K$ style word vectors for the target task OfficeHome~\cite{OfficeHomedataset} (65 classes) with respect to the design choices, \textit{Sequential} or \textit{Parallel}.}
    \vspace{-3.5mm}
    \label{fig:supp_design}
\end{figure}

\setcounter{figure}{0}
\setcounter{table}{0}
\renewcommand\thefigure{B\arabic{figure}}
\renewcommand\thetable{B\arabic{table}}

\begin{figure*}[t!]
    \centering
    \includegraphics[width=\textwidth]{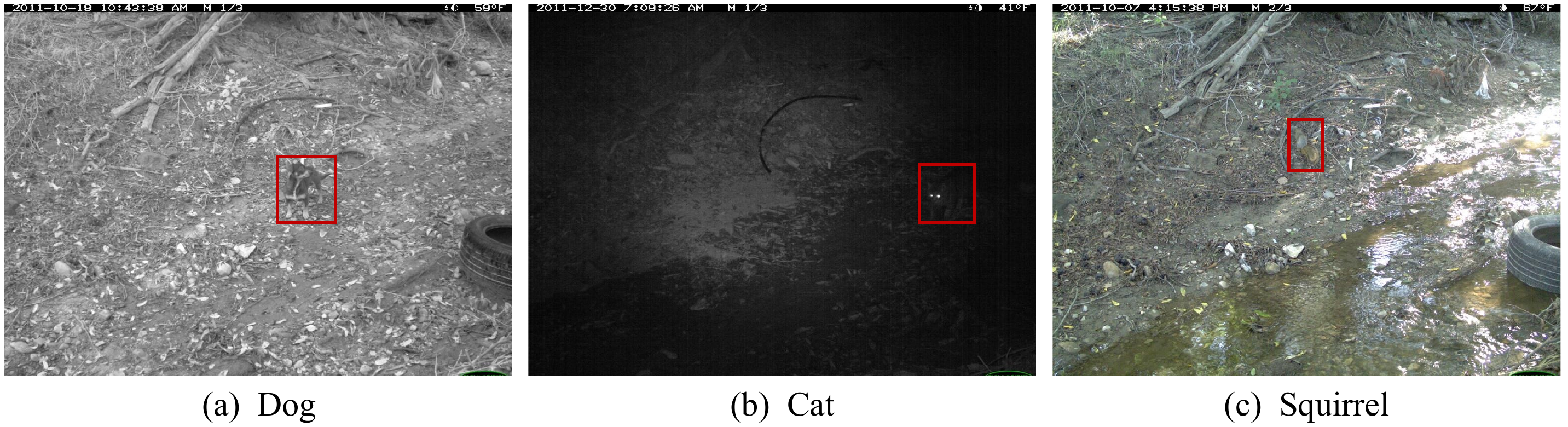}
    \vspace{-7mm}
    \caption{Several examples from the Terra Incognita~\cite{TerraDataset} dataset.
    We visualize class entities using red bounding boxes, since they are not easily recognizable due to their small sizes and complex background scenes.}
    \label{fig:supp_terra_dataset}
\end{figure*}

\begin{table*}[!t]
    \centering
    \resizebox{\textwidth}{!}{
        \begin{tabular}{lccccccc|c}
        \Xhline{2\arrayrulewidth}
        \multicolumn{1}{c}{}
        & \multicolumn{2}{c}{Configuration}
        & \multicolumn{1}{c}{}
        & \multicolumn{5}{c}{Accuracy (\%)}
        \\
        \cline{2-3}
        \cline{5-9}
        \vspace{-0.8mm}
        & Source & Domain & \;
        & & & &
        &
        \\
        Method
        & Domain & Description &
        & \normalsize{L}\small{ocation100} & \normalsize{L}\small{ocation38} & \normalsize{L}\small{ocation43} & \normalsize{L}\small{ocation46} & \normalsize{A}\small{vg.}
        \\
        \hline
        \multicolumn{9}{c}{\textit{ResNet-50~\cite{resnet} with pre-trained weights on ImageNet~\cite{deng2009imagenet}}}
        \\
        \hhline{-|-|-|-|-|-|-|-|-|}
            SelfReg~\cite{SelfReg}
            & \ding{51} & \textbf{--} &
            & 48.8\scriptsize{$\pm{0.9}$} & 41.3\scriptsize{$\pm{1.8}$} & 57.3\scriptsize{$\pm{0.7}$} & \textbf{40.6}\scriptsize{$\pm{0.9}$} & 47.0
        \\
            GVRT~\cite{Min2022Grounding}
            & \ding{51} & \textbf{--} &
            &  \textbf{53.9}\scriptsize{$\pm{1.3}$} & \textbf{41.8}\scriptsize{$\pm{1.2}$} & \textbf{58.2}\scriptsize{$\pm{0.9}$} & 38.0\scriptsize{$\pm{0.6}$} & \textbf{48.0}
        \\
        \hhline{-|-|-|-|-|-|-|-|-|}
        \multicolumn{9}{c}{\textit{ResNet-50~\cite{resnet} with pre-trained weights from CLIP~\cite{radford2021clip}}}
        \\
        \hhline{-|-|-|-|-|-|-|-|-|}
            ZS-CLIP (C)~\cite{radford2021clip}
            & \textbf{--} & \textbf{--} &
            & 8.4\scriptsize{$\pm{0.0}$} & 13.7\scriptsize{$\pm{0.0}$} & 32.5\scriptsize{$\pm{0.0}$} & 23.3\scriptsize{$\pm{0.0}$} & 19.5
        \\
            ZS-CLIP (PC)~\cite{radford2021clip}
            & \textbf{--} & \ding{51} &
            & 9.9\scriptsize{$\pm{0.0}$} & 28.3\scriptsize{$\pm{0.0}$} & 32.9\scriptsize{$\pm{0.0}$} & 24.0\scriptsize{$\pm{0.0}$} & 23.8
        \\
            \cellcolor{gray!9.0}\textbf{PromptStyler}\!\!
            & \cellcolor{gray!9.0}\textbf{--} 
            & \cellcolor{gray!9.0}\textbf{--} 
            & \cellcolor{gray!9.0} 
            & \cellcolor{gray!9.0}\textbf{13.8}\scriptsize{$\pm{1.7}$} & \cellcolor{gray!9.0}\textbf{39.8}\scriptsize{$\pm{1.3}$} & \cellcolor{gray!9.0}\textbf{38.0}\scriptsize{$\pm{0.4}$} & \cellcolor{gray!9.0}\textbf{30.3}\scriptsize{$\pm{0.3}$} & \cellcolor{gray!9.0}\textbf{30.5}
        \\
        \Xhline{2\arrayrulewidth}
    \end{tabular}}
    \vspace{-2mm}
    \caption{Top-1 classification accuracy on the Terra Incognita~\cite{TerraDataset} dataset.
    Compared with existing domain generalization methods which utilize source domain data, zero-shot methods using CLIP~\cite{radford2021clip} show unsatisfactory results on this dataset.}
    \label{table:supp_terra_limitation}
    \vspace{-1mm}
\end{table*}

\subsection{Empirical justification of our design choice}
\label{supp:subsection_design}
As described in Section~\redcolornumber{3.1} of the main paper,
there are two possible design choices for learning $K$ style word vectors: 
(1) learning each style word vector $\mathbf{s}_{i}$ in a sequential manner, or (2) learning all style word vectors $\{\mathbf{s}_{i}\}^{K}_{i=1}$ in a parallel manner.
We choose the former mainly due to its much less memory overhead.
As shown in Figure~\ref{fig:supp_design}, we could sequentially learn $\sim$$100$ style word vectors with $\sim$$4.2$ GB memory usage.
However, it is not possible to learn more than $21$ style word vectors in a parallel manner using a single RTX 3090 GPU (24 GB Memory) due to its large 
memory overhead.
In detail, learning $20$ and $21$ style word vectors takes $22.4$ GB and $23.5$ GB, respectively.
The large memory overhead caused by the parallel learning design substantially limits the number of learnable style word vectors. 

To be specific, PromptStyler with the parallel learning design needs to generate $K$ style features, $KN$ style-content features, and $N$ content features for learning $K$ style word vectors at the same time; 
these features are used to compute the style diversity loss $\mathcal{L}_{\mathrm{style}}$ and the content consistency loss $\mathcal{L}_{\mathrm{content}}$ for learning all the style word vectors in a parallel manner.
Note that the large memory overhead is mainly caused by the $KN$ style-content features.
Suppose we want to learn $80$ style word vectors for the target task OfficeHome~\cite{OfficeHomedataset} (65 classes).
Then, we need to synthesize $5200(=80 \times 65)$ style-content features.
Even worse, we need to generate $27600(=80 \times 345)$ style-content features for the target task DomainNet~\cite{DomainNetdataset} ($345$ classes).
On the other hand, PromptStyler with the sequential learning design only requires $i$ style features, $N$ style-content features, and $N$ content features for learning $i$-th style word vector, where $1 \leq i \leq K$.
For scalability, we chose the sequential learning design since it could handle a lot of learnable style word vectors and numerous classes in the target task.
\section{Analyses on Terra Incognita}
\label{supp:supp_2}
\vspace{-2mm}
As described in Section~\redcolornumber{5} of the main paper, the quality of the latent space constructed by a large-scale pre-trained model significantly affects the effectiveness of PromptStyler.
To be specific, the proposed method depends on the quality of the joint vision-language space constructed by CLIP~\cite{radford2021clip}.
Although our method achieves state-of-the-art results on PACS~\cite{PACSdataset}, VLCS~\cite{VLCSdataset}, OfficeHome~\cite{OfficeHomedataset}, and DomainNet~\cite{DomainNetdataset}, 
its performance on Terra Incognita~\cite{TerraDataset} is not satisfactory.
This section provides more analyses on the dataset.

Table~\ref{table:supp_terra_limitation} shows that PromptStyler outperforms zero-shot CLIP~\cite{radford2021clip} for all domains in the Terra Incognita dataset~\cite{TerraDataset}. 
However, its accuracy on this dataset is lower compared with existing domain generalization methods~\cite{Min2022Grounding,SelfReg} which utilize several images from the dataset as their source domain data.
This unsatisfactory result might be due to the low accuracy of CLIP on the dataset.
We suspect that images in the Terra Incognita dataset (Fig.~\ref{fig:supp_terra_dataset}) might be significantly different from the domains that CLIP has observed.
The distribution shifts between CLIP training dataset and the Terra Incognita dataset might be extreme, and thus such distribution shifts could not be entirely covered by our method
which exploits CLIP latent space.
We hope this issue could be alleviated with the development of large-scale models.
\setcounter{figure}{0}
\setcounter{table}{0}
\renewcommand\thefigure{C\arabic{figure}}
\renewcommand\thetable{C\arabic{table}}

\begin{table*}[!t]
    \centering
    \resizebox{\textwidth}{!}{
        \begin{tabular}{lccccccc|c}
        \Xhline{2\arrayrulewidth}
        \multicolumn{1}{c}{}
        & \multicolumn{2}{c}{Configuration}
        & \multicolumn{1}{c}{}
        & \multicolumn{5}{c}{Accuracy (\%)}
        \\
        \cline{2-3}
        \cline{5-9}

        \vspace{-0.8mm}
        & Source & Domain & \;
        & & & &
        &
        \\
        Method 
        & Domain & Description &
        & \normalsize{A}\small{rt Painting} & \normalsize{C}\small{artoon} & \normalsize{P}\small{hoto} & \normalsize{S}\small{ketch} 
        & \normalsize{A}\small{vg.}
        \\
        \hline
        \multicolumn{9}{c}{\textit{ResNet-50~\cite{resnet} with pre-trained weights on ImageNet~\cite{deng2009imagenet}}}
        \\
        \hhline{-|-|-|-|-|-|-|-|-|}
            GVRT~\cite{Min2022Grounding}
            & \ding{51} & \textbf{--} &  
            & \textbf{87.9}\scriptsize{$\pm{0.3}$}
            & 78.4\scriptsize{$\pm{1.0}$} 
            & \textbf{98.2}\scriptsize{$\pm{0.1}$} 
            & 75.7\scriptsize{$\pm{0.4}$} 
            & 85.1
        \\
            SelfReg~\cite{SelfReg}
            & \ding{51} & \textbf{--} &  
            & \textbf{87.9}\scriptsize{$\pm{1.0}$}
            & \textbf{79.4}\scriptsize{$\pm{1.4}$} 
            & 96.8\scriptsize{$\pm{0.7}$} 
            & \textbf{78.3}\scriptsize{$\pm{1.2}$} 
            & \textbf{85.6}
        \\
        \hhline{-|-|-|-|-|-|-|-|-|}
        \multicolumn{9}{c}{\textit{ResNet-50~\cite{resnet} with pre-trained weights from CLIP~\cite{radford2021clip}}}
        \\
        \hhline{-|-|-|-|-|-|-|-|-|}
            ZS-CLIP (C)~\cite{radford2021clip}
            & \textbf{--} & \textbf{--} &  
            & 88.9\scriptsize{$\pm{0.0}$}
            & 94.4\scriptsize{$\pm{0.0}$} 
            & 99.3\scriptsize{$\pm{0.0}$} 
            & 79.8\scriptsize{$\pm{0.0}$} 
            & 90.6
        \\
            ZS-CLIP (PC)~\cite{radford2021clip}
            & \textbf{--} & \ding{51} &  
            & 90.8\scriptsize{$\pm{0.0}$}
            & 93.3\scriptsize{$\pm{0.0}$} 
            & \textbf{99.4}\scriptsize{$\pm{0.0}$} 
            & 79.3\scriptsize{$\pm{0.0}$} 
            & 90.7
        \\
            \cellcolor{gray!9.0}\textbf{PromptStyler}\;\;
            & \cellcolor{gray!9.0}\textbf{--} 
            & \cellcolor{gray!9.0}\textbf{--} 
            & \cellcolor{gray!9.0}  
            & \cellcolor{gray!9.0}\textbf{93.7}\scriptsize{$\pm{0.1}$} 
            & \cellcolor{gray!9.0}\textbf{94.7}\scriptsize{$\pm{0.2}$}  
            & \cellcolor{gray!9.0}\textbf{99.4}\scriptsize{$\pm{0.0}$}  
            & \cellcolor{gray!9.0}\textbf{84.9}\scriptsize{$\pm{0.1}$}  
            & \cellcolor{gray!9.0}\textbf{93.2}
        \\
        \hhline{-|-|-|-|-|-|-|-|-|}
        \multicolumn{9}{c}{\textit{ViT-B\,/\,16~\cite{dosovitskiy2021an} with pre-trained weights from CLIP~\cite{radford2021clip}}}
        \\
        \hhline{-|-|-|-|-|-|-|-|-|}
            ZS-CLIP (C)~\cite{radford2021clip}
            & \textbf{--} & \textbf{--} &  
            & 96.4\scriptsize{$\pm{0.0}$}
            & 98.9\scriptsize{$\pm{0.0}$} 
            & \textbf{99.9}\scriptsize{$\pm{0.0}$} 
            & 87.7\scriptsize{$\pm{0.0}$} 
            & 95.7
        \\
            ZS-CLIP (PC)~\cite{radford2021clip}
            & \textbf{--} & \ding{51} &  
            & 97.2\scriptsize{$\pm{0.0}$}
            & \textbf{99.1}\scriptsize{$\pm{0.0}$} 
            & \textbf{99.9}\scriptsize{$\pm{0.0}$} 
            & 88.2\scriptsize{$\pm{0.0}$} 
            & 96.1
        \\
            \cellcolor{gray!9.0}\textbf{PromptStyler}
            & \cellcolor{gray!9.0}\textbf{--} & \cellcolor{gray!9.0}\textbf{--} & \cellcolor{gray!9.0}  
            & \cellcolor{gray!9.0}\textbf{97.6}\scriptsize{$\pm{0.1}$}
            & \cellcolor{gray!9.0}\textbf{99.1}\scriptsize{$\pm{0.1}$} 
            & \cellcolor{gray!9.0}\textbf{99.9}\scriptsize{$\pm{0.0}$} 
            & \cellcolor{gray!9.0}\textbf{92.3}\scriptsize{$\pm{0.3}$} 
            & \cellcolor{gray!9.0}\textbf{97.2}
        \\
        \hhline{-|-|-|-|-|-|-|-|-|}
        \multicolumn{9}{c}{\textit{ViT-L\,/\,14~\cite{dosovitskiy2021an} with pre-trained weights from CLIP~\cite{radford2021clip}}}
        \\
        \hhline{-|-|-|-|-|-|-|-|-|}
            ZS-CLIP (C)~\cite{radford2021clip}
            & \textbf{--} & \textbf{--} &  
            & 97.2\scriptsize{$\pm{0.0}$}
            & 99.5\scriptsize{$\pm{0.0}$} 
            & 99.9\scriptsize{$\pm{0.0}$} 
            & 93.8\scriptsize{$\pm{0.0}$} 
            & 97.6
        \\
            ZS-CLIP (PC)~\cite{radford2021clip}
            & \textbf{--} & \ding{51} &  
            & 99.0\scriptsize{$\pm{0.0}$}
            & \textbf{99.7}\scriptsize{$\pm{0.0}$} 
            & 99.9\scriptsize{$\pm{0.0}$} 
            & \textbf{95.5}\scriptsize{$\pm{0.0}$} 
            & 98.5
        \\
            \cellcolor{gray!9.0}\textbf{PromptStyler}
            & \cellcolor{gray!9.0}\textbf{--} & \cellcolor{gray!9.0}\textbf{--} & \cellcolor{gray!9.0}  
            & \cellcolor{gray!9.0}\textbf{99.1}\scriptsize{$\pm{0.0}$}
            & \cellcolor{gray!9.0}\textbf{99.7}\scriptsize{$\pm{0.0}$} 
            & \cellcolor{gray!9.0}\textbf{100.0}\scriptsize{$\pm{0.0}$} 
            & \cellcolor{gray!9.0}\textbf{95.5}\scriptsize{$\pm{0.1}$} 
            & \cellcolor{gray!9.0}\textbf{98.6}
        \\
        \Xhline{2\arrayrulewidth}
    \end{tabular}}
    \vspace{-2mm}
    \caption{Comparison with state-of-the-art domain generalization methods in terms of per-domain top-1 classification accuracy on PACS~\cite{PACSdataset}.
    We repeat each experiment using three different seeds, and report average accuracies with standard errors.
    ZS-CLIP (C) denotes zero-shot CLIP using ``[class]" as its text prompt, and ZS-CLIP (PC) indicates zero-shot CLIP using ``a photo of a [class]" as its text prompt.
    Note that \mbox{PromptStyler} does not use any source domain data and domain descriptions.}
    \vspace{3mm}
    \label{table:per_domain_pacs}
\end{table*}


\begin{table*}[!t]
    \centering
    \resizebox{\textwidth}{!}{
        \begin{tabular}{lccccccc|c}
        \Xhline{2\arrayrulewidth}
        \multicolumn{1}{c}{}
        & \multicolumn{2}{c}{Configuration}
        & \multicolumn{1}{c}{}
        & \multicolumn{5}{c}{Accuracy (\%)}
        \\
        \cline{2-3}
        \cline{5-9}

        \vspace{-0.8mm}
        & Source & Domain & \;
        & & & &
        &
        \\
        Method 
        & Domain & Description &
        & \normalsize{C}\small{altech} & \normalsize{L}\small{abelMe} & \normalsize{S}\small{UN09} & \normalsize{V}\small{OC2007} 
        & \normalsize{A}\small{vg.}
        \\
        \hline
        \multicolumn{9}{c}{\textit{ResNet-50~\cite{resnet} with pre-trained weights on ImageNet~\cite{deng2009imagenet}}}
        \\
        \hhline{-|-|-|-|-|-|-|-|-|}
            SelfReg~\cite{SelfReg}
            & \ding{51} & \textbf{--} &  
            & 96.7\scriptsize{$\pm{0.4}$}
            & \textbf{65.2}\scriptsize{$\pm{1.2}$} 
            & 73.1\scriptsize{$\pm{1.3}$} 
            & 76.2\scriptsize{$\pm{0.7}$} 
            & 77.8
        \\
            GVRT~\cite{Min2022Grounding}
            & \ding{51} & \textbf{--} &  
            & \textbf{98.8}\scriptsize{$\pm{0.1}$}
            & 64.0\scriptsize{$\pm{0.3}$} 
            & \textbf{75.2}\scriptsize{$\pm{0.5}$} 
            & \textbf{77.9}\scriptsize{$\pm{1.0}$} 
            & \textbf{79.0}
        \\
        \hhline{-|-|-|-|-|-|-|-|-|}
        \multicolumn{9}{c}{\textit{ResNet-50~\cite{resnet} with pre-trained weights from CLIP~\cite{radford2021clip}}}
        \\
        \hhline{-|-|-|-|-|-|-|-|-|}
            ZS-CLIP (C)~\cite{radford2021clip}
            & \textbf{--} & \textbf{--} &  
            & 99.2\scriptsize{$\pm{0.0}$}
            & 62.4\scriptsize{$\pm{0.0}$} 
            & 69.0\scriptsize{$\pm{0.0}$} 
            & 73.5\scriptsize{$\pm{0.0}$} 
            & 76.0
        \\
            ZS-CLIP (PC)~\cite{radford2021clip}
            & \textbf{--} & \ding{51} &  
            & 99.4\scriptsize{$\pm{0.0}$}
            & 65.0\scriptsize{$\pm{0.0}$} 
            & 71.7\scriptsize{$\pm{0.0}$} 
            & 84.2\scriptsize{$\pm{0.0}$} 
            & 80.1
        \\
            \cellcolor{gray!9.0}\textbf{PromptStyler}\;\;
            & \cellcolor{gray!9.0}\textbf{--} 
            & \cellcolor{gray!9.0}\textbf{--} 
            & \cellcolor{gray!9.0} 
            & \cellcolor{gray!9.0}\textbf{99.5}\scriptsize{$\pm{0.0}$}
            & \cellcolor{gray!9.0}\textbf{71.2}\scriptsize{$\pm{0.2}$}
            & \cellcolor{gray!9.0}\textbf{72.0}\scriptsize{$\pm{0.0}$}
            & \cellcolor{gray!9.0}\textbf{86.5}\scriptsize{$\pm{0.3}$}
            & \cellcolor{gray!9.0}\textbf{82.3}
        \\
        \hhline{-|-|-|-|-|-|-|-|-|}
        \multicolumn{9}{c}{\textit{ViT-B\,/\,16~\cite{dosovitskiy2021an} with pre-trained weights from CLIP~\cite{radford2021clip}}}
        \\
        \hhline{-|-|-|-|-|-|-|-|-|}
            ZS-CLIP (C)~\cite{radford2021clip}
            & \textbf{--} & \textbf{--} &  
            & 99.7\scriptsize{$\pm{0.0}$}
            & 61.8\scriptsize{$\pm{0.0}$} 
            & 70.1\scriptsize{$\pm{0.0}$} 
            & 73.9\scriptsize{$\pm{0.0}$} 
            & 76.4
        \\
            ZS-CLIP (PC)~\cite{radford2021clip}
            & \textbf{--} & \ding{51} &  
            & \textbf{99.9}\scriptsize{$\pm{0.0}$}
            & 68.9\scriptsize{$\pm{0.0}$} 
            & \textbf{74.8}\scriptsize{$\pm{0.0}$} 
            & 85.9\scriptsize{$\pm{0.0}$} 
            & 82.4
        \\
            \cellcolor{gray!9.0}\textbf{PromptStyler}
            & \cellcolor{gray!9.0}\textbf{--} & \cellcolor{gray!9.0}\textbf{--} & \cellcolor{gray!9.0}  
            & \cellcolor{gray!9.0}\textbf{99.9}\scriptsize{$\pm{0.0}$}
            & \cellcolor{gray!9.0}\textbf{71.5}\scriptsize{$\pm{0.3}$} 
            & \cellcolor{gray!9.0}73.9\scriptsize{$\pm{0.2}$} 
            & \cellcolor{gray!9.0}\textbf{86.3}\scriptsize{$\pm{0.1}$} 
            & \cellcolor{gray!9.0}\textbf{82.9}
        \\
        \hhline{-|-|-|-|-|-|-|-|-|}
        \multicolumn{9}{c}{\textit{ViT-L\,/\,14~\cite{dosovitskiy2021an} with pre-trained weights from CLIP~\cite{radford2021clip}}}
        \\
        \hhline{-|-|-|-|-|-|-|-|-|}
            ZS-CLIP (C)~\cite{radford2021clip}
            & \textbf{--} & \textbf{--} &  
            & \textbf{99.9}\scriptsize{$\pm{0.0}$}
            & 59.3\scriptsize{$\pm{0.0}$} 
            & 71.0\scriptsize{$\pm{0.0}$} 
            & 79.9\scriptsize{$\pm{0.0}$} 
            & 77.5
        \\
            ZS-CLIP (PC)~\cite{radford2021clip}
            & \textbf{--} & \ding{51} &  
            & \textbf{99.9}\scriptsize{$\pm{0.0}$}
            & 70.9\scriptsize{$\pm{0.0}$} 
            & \textbf{72.9}\scriptsize{$\pm{0.0}$} 
            & 86.0\scriptsize{$\pm{0.0}$} 
            & \textbf{82.4}
        \\
            \cellcolor{gray!9.0}\textbf{PromptStyler}
            & \cellcolor{gray!9.0}\textbf{--} & \cellcolor{gray!9.0}\textbf{--} & \cellcolor{gray!9.0}  
            & \cellcolor{gray!9.0}\textbf{99.9}\scriptsize{$\pm{0.0}$}
            & \cellcolor{gray!9.0}\textbf{71.1}\scriptsize{$\pm{0.7}$} 
            & \cellcolor{gray!9.0}71.8\scriptsize{$\pm{1.0}$} 
            & \cellcolor{gray!9.0}\textbf{86.8}\scriptsize{$\pm{0.0}$} 
            & \cellcolor{gray!9.0}\textbf{82.4}
        \\
        \Xhline{2\arrayrulewidth}
    \end{tabular}}
    \vspace{-2mm}
    \caption{Comparison with state-of-the-art domain generalization methods in terms of per-domain top-1 classification accuracy on VLCS~\cite{VLCSdataset}.
    We repeat each experiment using three different seeds, and report average accuracies with standard errors.
    ZS-CLIP (C) denotes zero-shot CLIP using ``[class]" as its text prompt, and ZS-CLIP (PC) indicates zero-shot CLIP using ``a photo of a [class]" as its text prompt.
    Note that \mbox{PromptStyler} does not use any source domain data and domain descriptions.}
    \label{table:per_domain_vlcs}
\end{table*}


\begin{table*}[!t]
    \centering
    \resizebox{\textwidth}{!}{
        \begin{tabular}{lccccccc|c}
        \Xhline{2\arrayrulewidth}
        \multicolumn{1}{c}{}
        & \multicolumn{2}{c}{Configuration}
        & \multicolumn{1}{c}{}
        & \multicolumn{5}{c}{Accuracy (\%)}
        \\
        \cline{2-3}
        \cline{5-9}

        \vspace{-0.8mm}
        & Source & Domain & \;
        & & & &
        &
        \\
        Method 
        & Domain & Description &
        & \normalsize{A}\small{rt} & \normalsize{C}\small{lipart} & \normalsize{P}\small{roduct} & \normalsize{R}\small{eal World} 
        & \normalsize{A}\small{vg.}
        \\
        \hline
        \multicolumn{9}{c}{\textit{ResNet-50~\cite{resnet} with pre-trained weights on ImageNet~\cite{deng2009imagenet}}}
        \\
        \hhline{-|-|-|-|-|-|-|-|-|}
            SelfReg~\cite{SelfReg}
            & \ding{51} & \textbf{--} &  
            & 63.6\scriptsize{$\pm{1.4}$}
            & 53.1\scriptsize{$\pm{1.0}$} 
            & 76.9\scriptsize{$\pm{0.4}$} 
            & 78.1\scriptsize{$\pm{0.4}$} 
            & 67.9
        \\
            GVRT~\cite{Min2022Grounding}
            & \ding{51} & \textbf{--} &  
            & \textbf{66.3}\scriptsize{$\pm{0.1}$}
            & \textbf{55.8}\scriptsize{$\pm{0.4}$} 
            & \textbf{78.2}\scriptsize{$\pm{0.4}$} 
            & \textbf{80.4}\scriptsize{$\pm{0.2}$} 
            & \textbf{70.1}
        \\
        \hhline{-|-|-|-|-|-|-|-|-|}
        \multicolumn{9}{c}{\textit{ResNet-50~\cite{resnet} with pre-trained weights from CLIP~\cite{radford2021clip}}}
        \\
        \hhline{-|-|-|-|-|-|-|-|-|}
            ZS-CLIP (C)~\cite{radford2021clip}
            & \textbf{--} & \textbf{--} &  
            & 69.9\scriptsize{$\pm{0.0}$}
            & 46.8\scriptsize{$\pm{0.0}$} 
            & 77.7\scriptsize{$\pm{0.0}$} 
            & 79.8\scriptsize{$\pm{0.0}$} 
            & 68.6
        \\
            ZS-CLIP (PC)~\cite{radford2021clip}
            & \textbf{--} & \ding{51} &  
            & 71.7\scriptsize{$\pm{0.0}$}
            & 52.0\scriptsize{$\pm{0.0}$} 
            & 81.6\scriptsize{$\pm{0.0}$} 
            & 82.6\scriptsize{$\pm{0.0}$} 
            & 72.0
        \\
            \cellcolor{gray!9.0}\textbf{PromptStyler}\;\;
            & \cellcolor{gray!9.0}\textbf{--} 
            & \cellcolor{gray!9.0}\textbf{--} 
            & \cellcolor{gray!9.0} 
            & \cellcolor{gray!9.0}\textbf{73.4}\scriptsize{$\pm{0.1}$}
            & \cellcolor{gray!9.0}\textbf{52.4}\scriptsize{$\pm{0.2}$} 
            & \cellcolor{gray!9.0}\textbf{84.3}\scriptsize{$\pm{0.1}$} 
            & \cellcolor{gray!9.0}\textbf{84.1}\scriptsize{$\pm{0.1}$} 
            & \cellcolor{gray!9.0}\textbf{73.6}
        \\
        \hhline{-|-|-|-|-|-|-|-|-|}
        \multicolumn{9}{c}{\textit{ViT-B\,/\,16~\cite{dosovitskiy2021an} with pre-trained weights from CLIP~\cite{radford2021clip}}}
        \\
        \hhline{-|-|-|-|-|-|-|-|-|}
            ZS-CLIP (C)~\cite{radford2021clip}
            & \textbf{--} & \textbf{--} &  
            & 80.7\scriptsize{$\pm{0.0}$}
            & 64.6\scriptsize{$\pm{0.0}$} 
            & 86.3\scriptsize{$\pm{0.0}$} 
            & 88.0\scriptsize{$\pm{0.0}$} 
            & 79.9
        \\
            ZS-CLIP (PC)~\cite{radford2021clip}
            & \textbf{--} & \ding{51} &  
            & 82.7\scriptsize{$\pm{0.0}$}
            & 67.6\scriptsize{$\pm{0.0}$} 
            & 89.2\scriptsize{$\pm{0.0}$} 
            & 89.7\scriptsize{$\pm{0.0}$} 
            & 82.3
        \\
            \cellcolor{gray!9.0}\textbf{PromptStyler}
            & \cellcolor{gray!9.0}\textbf{--} & \cellcolor{gray!9.0}\textbf{--} & \cellcolor{gray!9.0}  
            & \cellcolor{gray!9.0}\textbf{83.8}\scriptsize{$\pm{0.1}$}
            & \cellcolor{gray!9.0}\textbf{68.2}\scriptsize{$\pm{0.0}$} 
            & \cellcolor{gray!9.0}\textbf{91.6}\scriptsize{$\pm{0.1}$} 
            & \cellcolor{gray!9.0}\textbf{90.7}\scriptsize{$\pm{0.1}$} 
            & \cellcolor{gray!9.0}\textbf{83.6}
        \\
        \hhline{-|-|-|-|-|-|-|-|-|}
        \multicolumn{9}{c}{\textit{ViT-L\,/\,14~\cite{dosovitskiy2021an} with pre-trained weights from CLIP~\cite{radford2021clip}}}
        \\
        \hhline{-|-|-|-|-|-|-|-|-|}
            ZS-CLIP (C)~\cite{radford2021clip}
            & \textbf{--} & \textbf{--} &  
            & 86.2\scriptsize{$\pm{0.0}$}
            & 73.3\scriptsize{$\pm{0.0}$} 
            & 92.0\scriptsize{$\pm{0.0}$} 
            & 92.2\scriptsize{$\pm{0.0}$} 
            & 85.9
        \\
            ZS-CLIP (PC)~\cite{radford2021clip}
            & \textbf{--} & \ding{51} &  
            & 87.2\scriptsize{$\pm{0.0}$}
            & 73.8\scriptsize{$\pm{0.0}$} 
            & 93.0\scriptsize{$\pm{0.0}$} 
            & 93.4\scriptsize{$\pm{0.0}$} 
            & 86.9
        \\
            \cellcolor{gray!9.0}\textbf{PromptStyler}
            & \cellcolor{gray!9.0}\textbf{--} & \cellcolor{gray!9.0}\textbf{--} & \cellcolor{gray!9.0}  
            & \cellcolor{gray!9.0}\textbf{89.1}\scriptsize{$\pm{0.1}$}
            & \cellcolor{gray!9.0}\textbf{77.6}\scriptsize{$\pm{0.1}$} 
            & \cellcolor{gray!9.0}\textbf{94.8}\scriptsize{$\pm{0.1}$} 
            & \cellcolor{gray!9.0}\textbf{94.8}\scriptsize{$\pm{0.0}$} 
            & \cellcolor{gray!9.0}\textbf{89.1}
        \\ 
        \Xhline{2\arrayrulewidth}
    \end{tabular}}
    \vspace{-2mm}
    \caption{Comparison with state-of-the-art domain generalization methods in terms of per-domain top-1 classification accuracy on OfficeHome~\cite{OfficeHomedataset}.
    We repeat each experiment using three different seeds, and report average accuracies with standard errors.
    ZS-CLIP (C) denotes zero-shot CLIP using ``[class]" as its text prompt, and ZS-CLIP (PC) indicates zero-shot CLIP using ``a photo of a [class]" as its text prompt.
    Note that \mbox{PromptStyler} does not use any source domain data and domain descriptions.}
    \vspace{6mm}
    \label{table:per_domain_officehome}
\end{table*}


\begin{table*}[!t]
    \centering
    \resizebox{\textwidth}{!}{
        \begin{tabular}{lccccccccc|c}
        \Xhline{2\arrayrulewidth}
        \multicolumn{1}{c}{}
        & \multicolumn{2}{c}{Configuration}
        & \multicolumn{1}{c}{}
        & \multicolumn{7}{c}{Accuracy (\%)}
        \\
        \cline{2-3}
        \cline{5-11}

        \vspace{-0.8mm}
        & \!\!Source\!\! & \!\!Domain\!\! & 
        & & & & & &
        &
        \\
        Method 
        & \!\!Domain\!\! & \!\!Description\!\! &
        & \!\!\!\!\normalsize{C}\small{lipart}\!\!\!\! & \!\!\!\!\normalsize{I}\small{nfograph}\!\!\!\! & \!\!\!\!\normalsize{P}\small{ainting}\!\!\!\! & \!\!\!\!\normalsize{Q}\small{uickdraw}\!\!\!\! & \!\!\!\!\!\!\!\!\!\!\normalsize{R}\small{eal}\!\!\!\!\!\!\!\!\!\! & \!\!\!\!\normalsize{S}\small{ketch}\!\!\!\! 
        & \!\!\!\!\normalsize{A}\small{vg.}\!\!\!\!
        \\
        \hline
        \multicolumn{11}{c}{\textit{ResNet-50~\cite{resnet} with pre-trained weights on ImageNet~\cite{deng2009imagenet}}}
        \\
        \hhline{-|-|-|-|-|-|-|-|-|-|-|}
            SelfReg~\cite{SelfReg}
            & \ding{51} & \textbf{--} &  
            & 60.7\scriptsize{$\pm{0.1}$}
            & \textbf{21.6}\scriptsize{$\pm{0.1}$} 
            & 49.4\scriptsize{$\pm{0.2}$} 
            & 12.7\scriptsize{$\pm{0.1}$} 
            & 60.7\scriptsize{$\pm{0.1}$} 
            & 51.7\scriptsize{$\pm{0.1}$} 
            & 42.8
        \\
            GVRT~\cite{Min2022Grounding}
            & \ding{51} & \textbf{--} &  
            & \textbf{62.4}\scriptsize{$\pm{0.4}$}
            & 21.0\scriptsize{$\pm{0.0}$} 
            & \textbf{50.5}\scriptsize{$\pm{0.4}$} 
            & \textbf{13.8}\scriptsize{$\pm{0.3}$} 
            & \textbf{64.6}\scriptsize{$\pm{0.4}$} 
            & \textbf{52.4}\scriptsize{$\pm{0.2}$} 
            & \textbf{44.1}
        \\
        \hhline{-|-|-|-|-|-|-|-|-|-|-|}
        \multicolumn{11}{c}{\textit{ResNet-50~\cite{resnet} with pre-trained weights from CLIP~\cite{radford2021clip}}}
        \\
        \hhline{-|-|-|-|-|-|-|-|-|-|-|}
            ZS-CLIP (C)~\cite{radford2021clip}
            & \textbf{--} & \textbf{--} &  
            & 53.1\scriptsize{$\pm{0.0}$}
            & 39.2\scriptsize{$\pm{0.0}$} 
            & 52.7\scriptsize{$\pm{0.0}$} 
            & \textbf{6.3}\scriptsize{$\pm{0.0}$} 
            & 75.2\scriptsize{$\pm{0.0}$} 
            & 47.1\scriptsize{$\pm{0.0}$} 
            & 45.6
        \\
            ZS-CLIP (PC)~\cite{radford2021clip}
            & \textbf{--} & \ding{51} &  
            & 53.6\scriptsize{$\pm{0.0}$}
            & 39.6\scriptsize{$\pm{0.0}$} 
            & 53.4\scriptsize{$\pm{0.0}$} 
            & 5.9\scriptsize{$\pm{0.0}$} 
            & 76.6\scriptsize{$\pm{0.0}$} 
            & 48.0\scriptsize{$\pm{0.0}$} 
            & 46.2
        \\
            \cellcolor{gray!9.0}\textbf{PromptStyler}
            & \cellcolor{gray!9.0}\textbf{--} & \cellcolor{gray!9.0}\textbf{--} 
            & \cellcolor{gray!9.0}  
            & \cellcolor{gray!9.0}\textbf{57.9}\scriptsize{$\pm{0.0}$}
            & \cellcolor{gray!9.0}\textbf{44.3}\scriptsize{$\pm{0.0}$}
            & \cellcolor{gray!9.0}\textbf{57.3}\scriptsize{$\pm{0.0}$} 
            & \cellcolor{gray!9.0}6.1\scriptsize{$\pm{0.1}$} 
            & \cellcolor{gray!9.0}\textbf{79.5}\scriptsize{$\pm{0.0}$} 
            & \cellcolor{gray!9.0}\textbf{51.7}\scriptsize{$\pm{0.0}$} 
            & \cellcolor{gray!9.0}\textbf{49.5}
        \\
        \hhline{-|-|-|-|-|-|-|-|-|-|-|}
        \multicolumn{11}{c}{\textit{ViT-B\,/\,16~\cite{dosovitskiy2021an} with pre-trained weights from CLIP~\cite{radford2021clip}}}
        \\
        \hhline{-|-|-|-|-|-|-|-|-|-|-|}
            ZS-CLIP (C)~\cite{radford2021clip}
            & \textbf{--} & \textbf{--} &  
            & 70.7\scriptsize{$\pm{0.0}$}
            & 49.1\scriptsize{$\pm{0.0}$} 
            & 66.4\scriptsize{$\pm{0.0}$} 
            & \textbf{14.8}\scriptsize{$\pm{0.0}$} 
            & 82.7\scriptsize{$\pm{0.0}$} 
            & 63.1\scriptsize{$\pm{0.0}$} 
            & 57.8
        \\
            ZS-CLIP (PC)~\cite{radford2021clip}
            & \textbf{--} & \ding{51} &  
            & 71.0\scriptsize{$\pm{0.0}$}
            & 47.7\scriptsize{$\pm{0.0}$} 
            & 66.2\scriptsize{$\pm{0.0}$} 
            & 14.0\scriptsize{$\pm{0.0}$} 
            & 83.7\scriptsize{$\pm{0.0}$} 
            & 63.5\scriptsize{$\pm{0.0}$} 
            & 57.7
        \\
            \cellcolor{gray!9.0}\textbf{PromptStyler}
            & \cellcolor{gray!9.0}\textbf{--} & \cellcolor{gray!9.0}\textbf{--} & \cellcolor{gray!9.0} 
            & \cellcolor{gray!9.0}\textbf{73.1}\scriptsize{$\pm{0.0}$}
            & \cellcolor{gray!9.0}\textbf{50.9}\scriptsize{$\pm{0.0}$} 
            & \cellcolor{gray!9.0}\textbf{68.2}\scriptsize{$\pm{0.1}$} 
            & \cellcolor{gray!9.0}13.3\scriptsize{$\pm{0.1}$}
            & \cellcolor{gray!9.0}\textbf{85.4}\scriptsize{$\pm{0.0}$} 
            & \cellcolor{gray!9.0}\textbf{65.3}\scriptsize{$\pm{0.0}$} 
            & \cellcolor{gray!9.0}\textbf{59.4}
        \\
        \hhline{-|-|-|-|-|-|-|-|-|-|-|}
        \multicolumn{11}{c}{\textit{ViT-L\,/\,14~\cite{dosovitskiy2021an} with pre-trained weights from CLIP~\cite{radford2021clip}}}
        \\
        \hhline{-|-|-|-|-|-|-|-|-|-|-|}
            ZS-CLIP (C)~\cite{radford2021clip}
            & \textbf{--} & \textbf{--} &  
            & 78.2\scriptsize{$\pm{0.0}$}
            & 53.0\scriptsize{$\pm{0.0}$} 
            & 70.7\scriptsize{$\pm{0.0}$} 
            & 21.6\scriptsize{$\pm{0.0}$} 
            & 86.0\scriptsize{$\pm{0.0}$} 
            & 70.3\scriptsize{$\pm{0.0}$} 
            & 63.3
        \\
            ZS-CLIP (PC)~\cite{radford2021clip}
            & \textbf{--} & \ding{51} &  
            & 79.2\scriptsize{$\pm{0.0}$}
            & 52.4\scriptsize{$\pm{0.0}$} 
            & 71.3\scriptsize{$\pm{0.0}$} 
            & \textbf{22.5}\scriptsize{$\pm{0.0}$} 
            & 86.9\scriptsize{$\pm{0.0}$} 
            & 71.8\scriptsize{$\pm{0.0}$} 
            & 64.0
        \\
            \cellcolor{gray!9.0}\textbf{PromptStyler}
            & \cellcolor{gray!9.0}\textbf{--} & \cellcolor{gray!9.0}\textbf{--} & \cellcolor{gray!9.0} 
            & \cellcolor{gray!9.0}\textbf{80.7}\scriptsize{$\pm{0.0}$}
            & \cellcolor{gray!9.0}\textbf{55.6}\scriptsize{$\pm{0.1}$} 
            & \cellcolor{gray!9.0}\textbf{73.8}\scriptsize{$\pm{0.1}$} 
            & \cellcolor{gray!9.0}21.7\scriptsize{$\pm{0.0}$}
            & \cellcolor{gray!9.0}\textbf{88.2}\scriptsize{$\pm{0.0}$} 
            & \cellcolor{gray!9.0}\textbf{73.2}\scriptsize{$\pm{0.0}$} 
            & \cellcolor{gray!9.0}\textbf{65.5}
        \\
        \Xhline{2\arrayrulewidth}
    \end{tabular}}
    \vspace{-2mm}
    \caption{Comparison with state-of-the-art domain generalization methods in terms of per-domain top-1 classification accuracy on DomainNet~\cite{DomainNetdataset}.
    We repeat each experiment using three different seeds, and report average accuracies with standard errors.
    ZS-CLIP (C) denotes zero-shot CLIP using ``[class]" as its text prompt, and ZS-CLIP (PC) indicates zero-shot CLIP using ``a photo of a [class]" as its text prompt.
    Note that \mbox{PromptStyler} does not use any source domain data and domain descriptions.}
    \label{table:per_domain_domainnet}
\end{table*}


\section{Evaluation Results}
\label{supp:supp_3}

\noindent \textbf{Per-domain accuracy.}
As shown in Table~\ref{table:per_domain_pacs}--\ref{table:per_domain_domainnet},
we provide per-domain top-1 classification accuracy on domain generalization benchmarks including PACS~\cite{PACSdataset} (4 domains and 7 classes), VLCS~\cite{VLCSdataset} (4 domains and 5 classes), OfficeHome~\cite{OfficeHomedataset} (4 domains and 65 classes) and DomainNet~\cite{DomainNetdataset} (6 domains and 345 classes);
each accuracy is obtained by averaging results from experiments repeated using three different random seeds. 
Interestingly, compared with zero-shot CLIP~\cite{radford2021clip} which leverages a photo domain description (``a photo of a [class]"), our PromptStyler achieves similar or better results on photo domains, \eg, on the VLCS dataset which consists of 4 photo domains.
Note that the description has more domain-specific information and more detailed contexts compared with the na\"ive prompt (``[class]").

\begin{table}[!t]
    \centering
    \resizebox{\columnwidth}{!}{
        \begin{tabular}{lcccc|c}
        \Xhline{2\arrayrulewidth}
        & \multicolumn{5}{c}{\small{Accuracy (\%)}}
        \\
        \cline{2-6}
        \!Distribution
        & \!\small{P}\scriptsize{ACS}\! & \!\small{V}\scriptsize{LCS}\! & \!\small{O}\scriptsize{fficeHome}\!\! & \!\small{D}\scriptsize{omainNet}\! & \!\small{A}\scriptsize{vg.}\!
        \\
        \hline
            \!$\mathcal{U}(0.00,\,0.20)$
            & \!93.1\! & \!\textbf{82.6}\! & \!\textbf{73.8}\!\! & \!49.2\! & \!\textbf{74.7}\!
        \\
            \!\!$\mathcal{N}(0.00,\,0.20^{2})$
            & \!93.0\! & \!81.0\! & \!73.6\!\! & \!\textbf{49.5}\! & \!74.3\!
        \\
            \!\!$\mathcal{N}(0.20,\,0.02^{2})$
            & \!93.1\! & \!82.5\! & \!73.5\!\! & \!49.3\! & \!74.6\!
        \\
            \cellcolor{gray!9.0}\!\!$\mathcal{N}(0.00,\,0.02^{2})$
            & \cellcolor{gray!9.0}\!\textbf{93.2}\! & \cellcolor{gray!9.0}\!82.3\! & \cellcolor{gray!9.0}\!73.6\!\! & \cellcolor{gray!9.0}\!\textbf{49.5}\! & \cellcolor{gray!9.0}\!\textbf{74.7}\!
        \\
        \Xhline{2\arrayrulewidth}
    \end{tabular}}
    \vspace{-2mm}
    \caption{Effects of the distributions used for initializing style word vectors.
    Uniform or Normal distribution is used.}
    \vspace{-1mm}
    \label{table:supp_initialization_method}
\end{table}

\noindent \textbf{Different distributions for initializing style word vectors.} 
Following prompt learning methods~\cite{CoOp,CoCoOp}, we initialized learnable style word vectors using zero-mean Gaussian distribution with $0.02$ standard deviation.
To measure the effect of the used distribution for the initialization,
we also quantitatively evaluate PromptStyler using different distributions for initializing style word vectors.
As shown in Table~\ref{table:supp_initialization_method},
the proposed method also achieves similar results when initializing style word vectors using different distributions.
\setcounter{figure}{0}
\setcounter{table}{0}
\renewcommand\thefigure{D\arabic{figure}}
\renewcommand\thetable{D\arabic{table}}

\section{Discussion}
\label{supp:supp_4}
PromptStyler aims to improve model's generalization capability by simulating various distribution shifts in the latent space of a large-scale pre-trained model.
To achieve this goal, our method leverages a joint vision-language space where text features could effectively represent their relevant image features.
It does not mean that image and text features should be perfectly interchangeable in the joint vision-language space;
a recent study has demonstrated the modality gap phenomenon of this joint space~\cite{liang2022mind}.
However, thanks to the cross-modal transferability in the joint vision-language space~\cite{zhang2023diagnosing},
the proposed method could still be effective,
\ie, we could consider text features as proxies for image features while training a linear classifier (Fig.~\redcolornumber{3} of the main paper).

When our method is implemented with CLIP~\cite{radford2021clip} and we adopt ArcFace~\cite{ArcFace} as our classification loss $\mathcal{L}_{\mathrm{class}}$, there is another interesting interpretation of the proposed method.
As described in Section~\redcolornumber{A.1},
CLIP text encoder synthesizes classifier weights using class names for zero-shot inference and then it computes cosine similarity scores between the classifier weights and input image features.
Similarly, our method computes cosine similarity scores between classifier weights of the trained classifier (Fig.~\redcolornumber{3} of the main paper) and input image features.
From this perspective,
the proposed method improves the decision boundary of the synthesized classifier used in zero-shot CLIP by generating diverse style-content features and then training a linear classifier using the style-content features. 
In other words, 
the trained classifier could be considered as an improved version of the synthesized classifier used in zero-shot CLIP.

\end{document}